\documentclass[keeplastbox, letterpaper, 10 pt, conference]{ieeeconf}  

\IEEEoverridecommandlockouts                              

\makeatletter
\let\NAT@parse\undefined
\makeatother

\usepackage{graphics}  
\usepackage{graphicx}

\usepackage{bm}        
\usepackage{amsmath}   
\usepackage{amssymb}   

\usepackage{balance}

\usepackage{float}
\usepackage{makecell}
\usepackage[normalem]{ulem}
\usepackage{hyperref}
\usepackage{color}

\usepackage{amsfonts}
\usepackage{algorithm,algpseudocode}

\usepackage{gensymb}
\usepackage[tbtags]{mathtools}

\usepackage[pdftex,dvipsnames]{xcolor}

\usepackage{todonotes}
\usepackage{units}
\usepackage{booktabs}
\usepackage{tikz}
\usepackage{textcomp}

\usepackage[bottom]{footmisc}
\usepackage{etoolbox}

\DeclareMathOperator*{\argmin}{arg\,min}

\usepackage[comma,numbers,compress]{natbib}

\usepackage[firstpage=true]{background}

\SetBgContents{\textbf{Accepted final version.} In \textit{2020 IEEE/RSJ International Conference on Intelligent Robots and Systems (IROS)}}
\SetBgScale{1}
\SetBgColor{black}
\SetBgAngle{0}
\SetBgOpacity{1}
\SetBgPosition{current page.south}
\SetBgVshift{1cm}

\title{\LARGE \bf Category-Level 3D Non-Rigid Registration\\
	from Single-View RGB Images}
\author{Diego Rodriguez*, Florian Huber*, and Sven Behnke
	\thanks{*~Authors with equal contribution. 
		All authors are with the Autonomous	Intelligent Systems (AIS) Group, Computer Science Institute VI, University of Bonn, Germany {\tt\small rodriguez@ais.uni-bonn.de}}
	\thanks{This work was funded by grant BE 2556/16-1 (Research Unit FOR 2535 Anticipating Human Behavior) of the German Research Foundation (DFG) and an Amazon Research Award.}
}%

\begin{document}

\maketitle
\thispagestyle{empty}
\pagestyle{empty}

\begin{abstract}
	In this paper,
	we propose a novel approach to solve the 3D non-rigid registration problem from RGB images using Convolutional Neural Networks (CNNs).
	Our objective is to find a deformation field (typically used for transferring knowledge between instances, e.g., 
	grasping skills) that warps a given 3D canonical model into a novel instance observed by a single-view RGB image.
	This is done by training a CNN that infers a deformation field for the visible parts of the canonical model and by employing a learned shape (latent) space for inferring the deformations of the occluded parts.
	As result of the registration, 
	the observed model is reconstructed.
	Because our method does not need depth information, 
	it can register objects that are typically hard to perceive with RGB-D sensors, e.g. with transparent or shiny surfaces.
	Even without depth data, 
	our approach outperforms the Coherent Point Drift (CPD) registration method for the evaluated object categories.
\end{abstract}

\section{Introduction}
\label{sec:introduction}

Registering 3D objects in a non-rigid way is essential for many real world applications, including robot manipulation~\cite{Pavlichenko2019, Pavlichenko2018}, human body analysis~\cite{hasler2009statistical, allen2003space} and grasp planning~\cite{stouraitis, Rodriguez2018Learning}.
This registration plays a key role for transferring domain knowledge, 
such as the transfer of control poses for approaching and grasping objects.
This knowledge transfer is inspired by the observation that objects with similar shape and usage can be manipulated in an analogous manner---adapting knowledge from previous successful experiences to novel observed instances.
Following this idea, in this work, 
we register non-rigidly a canonical model towards novel instances of the same object category.

The 3D non-rigid registration problem is often addressed
based on RGB-D images or 3D scans~\cite{brown2007global,zollhofer2014real,Rodriguez2018b}.
However, 
several objects cannot be measured well by depth sensors,
e.g., transparent drinking bottles.
By registering objects from RGB images,
we attempt to address this issue.

The registration from a single RGB image is a challenging problem for several reasons.
First, 
3D deformations need to be calculated without depth information,
i.e., 
a mapping between 2D colored pixels and 3D deformations has to be established.
Second, 
from a single-view RGB image the object is not fully observable and occluded parts have to be reconstructed.
Third, 
this reconstruction problem is ambiguous; 
several plausible shapes can explain a single observation.

\begin{figure}
	\centering
	\includegraphics[width=\linewidth]{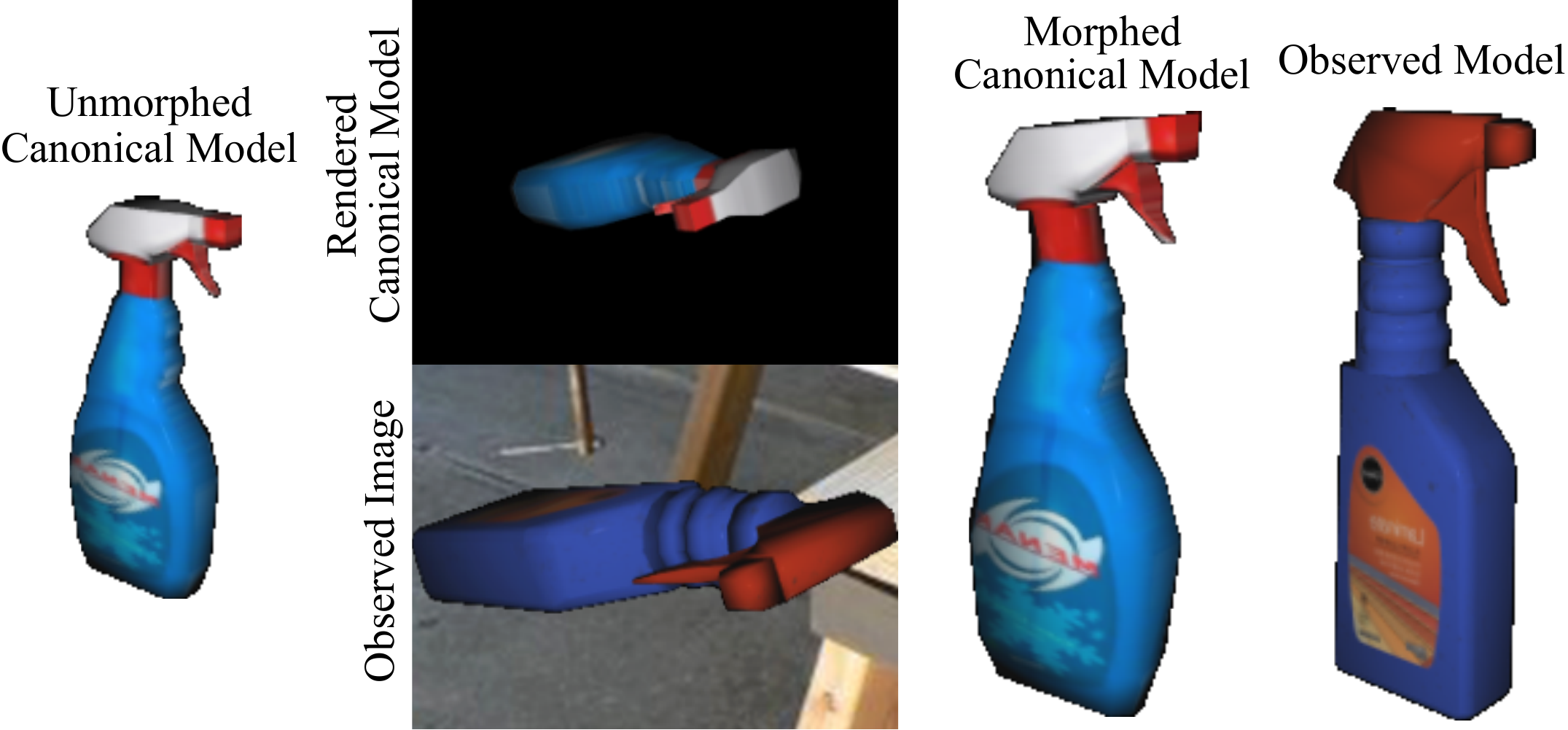}
	\caption{A canonical model of a category is deformed based on a single-view RGB image.
		The model is reconstructed without any depth information from partially observing the object. 
		Models are shown at the same scale to show the large deformation the canonical model has undergone.}
	\label{fig:teaser}
	\vspace*{-2ex}
\end{figure}

In this paper,
we propose a novel approach that is able to infer 3D non-rigid deformations from single-view RGB-only images of objects belonging to the same category\footnote{~Video: \href{https://www.ais.uni-bonn.de/videos/IROS_2020_Rodriguez}{www.ais.uni-bonn.de/videos/IROS\_2020\_Rodriguez}} (Fig.~\ref{fig:teaser}).
This is done by leveraging on realistic 3D object models.
A deep Convolutional Neural Network (CNN) is trained to infer deformations in the $x$, $y$ and $z$ axes, given the observed image of an object and a rendered image of the canonical model.
The training images are generated by rendering 3D realistic models and calculating ground truth deformations using the Coherent Point Drift (CPD) method.
By using a learned shape space that describes the typical geometrical variability of the object category, our approach is able to reconstruct the observed instances and provides deformations even for the non-observable parts.
Thus,
the reconstructed object exhibits a category-like shape.
The resulting deformation field allows to transfer category-level knowledge, e.g. grasping skills, to novel instances without further training.

The core contribution of this paper is to introduce a novel approach that registers non-rigidly a 3D model towards an object observed by a single RGB image.
In addition, 
we develop a method to generate a dataset suitable for non-rigid registration tasks.
This method is able to synthetically augment the number of instances of a category to overcome the scarcity of high-quality 3D textured models.

\section{Related Work}
\label{sec:related_work}

\subsection{Rendering for Deep Learning}
The use of synthetic data to generate training samples has been widely adopted in the deep learning community for object detection, semantic segmentation and pose estimation tasks~\cite{schwarz2020, tremblay2018falling, xiang2018posecnn, zhang2017physically}.
One of the first successful attempts was proposed by \citet{tobin2017domain},
who trained an object detection network using only synthetic data and were able to transfer the network to real-world applications.
More recently, \citet{tremblay2018deep} achieved state-of-the-art performance on 6-DoF pose estimation by using only synthetic data.
Our work is partially inspired by pose refinement approaches~\cite{deepIm}, 
that demonstrated that the rigid registration problem can be solved by the use of rendered images.

\subsection{Non-Rigid Registration}
Most of the approaches that address the 3D non-rigid registration problem assume the availability of full volumetric data.
These approaches mainly differ by the constraints imposed on how the points should move to match the observation including:
thin splines~\cite{allen2003space, brown2007global},
conformal maps~\cite{kim2011blended, zeng2010dense}, and coherent motion~\cite{myronenko2010point}.
However, 
these approaches do not perform well with partially observed objects.
In our previous work~\cite{Rodriguez2018a}, we incorporated a shape space of similar instances to address this issue.
Plausible geometries are inferred and a deformation field is provided for the non-observable parts, but depth information is required.
In contrast, 
the method presented in this paper registers a canonical model to novel object instances in a non-rigid manner without depth information---only a single-view RGB image of the observed object is required.
This is possible thanks to category-level prior knowledge of the objects represented in a CNN.

\subsection{Shape Reconstruction}
For 3D shape completion,
several approaches have been proposed, including radial basis functions~\cite{carr2001reconstruction}, surface primitives~\cite{kazhdan2013screened}, and Laplacian mesh optimization~\cite{nealen2006laplacian}. 
Data-driven approaches have been recently presented for 3D completion tasks. 
\citet{choy20163d} learn a mapping from object images to 3D shapes from a large collection of synthetic data by means of a Recurrent Neural Network (RNN).
In \cite{rock2015completing}, a 3D model is completed given a single depth image by transferring symmetries and surfaces from an exemplary database.
By integrating deep generative models with adversarially learned shape priors,
\citet{wu2018learning} are able to complete and reconstruct 3D shapes from single view images.

Similar to our work, 
the approaches proposed by \citet{wu2018learning} and \citet{choy20163d} also attempt to reconstruct a 3D model based on a RGB image.
While both approaches achieve good results when estimating shapes, 
they do not provide any estimate of the deformations between instances or any kind of registration.
Our approach, 
on the other hand, 
reconstructs the observed object and provides deformation fields enabling knowledge transfer to novel instances directly without the need for costly additional registrations posterior to the reconstruction.
\section{Background}
\label{sec:backgraound}
In our approach, 
we infer a deformation field that warps a canonical model into an observed instance from a single-view RGB image.
We train a deep CNN in a supervised manner to infer a deformation field from the observed input image.
This field is represented as three output feature maps representing the deformations in the $x$, $y$ and $z$ axes of the object.
The target data for training is calculated using the Coherent Point Drift~\cite{myronenko2010point} which is briefly introduced below.
To infer the deformation field of the non-observable parts,
a shape (latent) space of the object category is constructed (Sec.~\ref{sec:shape_space})~\cite{Rodriguez2018b,Rodriguez2018a}.

\subsection{Coherent Point Drift}
\label{sec:cpd}
For two sets of D-dimensional points,
$\mathbf{X}$ = $(\mathbf{x}_1 , ... , \mathbf{x}_N) ^T$ and $\mathbf{Y}$ = $(\mathbf{y}_1 , ... , \mathbf{y}_M) ^T$,
CPD outputs a deformation field mapping $\mathbf{Y}$ towards $\mathbf{X}$. 
For this purpose,
the points of $\mathbf{Y}$ are considered as centroids of a Gaussian Mixture Model (GMM) and the points of $\mathbf{X}$ are considered as samples drawn from these GMMs.
Equal isotropic covariances $\sigma^2$ and equal membership probabilities $P(m) = \frac{1}{M}$ are used for all GMM components. 
CPD outputs a deformed point set $\mathcal{T}$ as result of the maximization of the probability of the points from $\mathbf{X}$ being drawn from the GMMs, by moving the centroids $\mathbf{Y}$ in a coherent manner~\cite{motion_coherence}.
This optimization is formulated as an Expectation Maximization (EM) problem.

For non-rigid registration, the deformed point set $\mathcal{T}$ is described by the initial position of the points $\mathbf{Y}$ and a displacement function $v$:
\begin{equation} 
	\label{eq:cpd1}
	\mathcal{T}(\mathbf{Y},v) = \mathbf{Y} + v(\mathbf{Y}).
\end{equation}
For any D-dimensional set of points $\mathbf{Z} \in \mathbb{R}^{N\times D}$,
the displacement function $v$ is defined as:
\begin{equation} 
	\label{eq:cpd2}
	v(\mathbf{Z}) = \mathbf{G}(\mathbf{Y},\mathbf{Z}) \mathbf{W},
\end{equation}
where $\mathbf{W} \in \mathbb{R}^{N\times D}$ is a weight matrix,
which can be interpreted as a set of D-dimensional deformation vectors.
This matrix $\mathbf{W}$ is estimated in the M-step of the EM algorithm.
The coherent movement is controlled by the Gaussian kernel matrix $\mathbf{G}(\mathbf{Y},\mathbf{Z})$~\cite{myronenko2010point}, 
defined element-wise as:
\begin{equation} 
	\label{eq:G}
	g_{ij} = \textbf{G}(y_i,z_j) = \exp(-\frac{1}{ {2\beta}^2 } \left\lVert y_i - z_i \right\rVert ^ 2),
\end{equation}
where $\beta$ is a parameter that controls the interaction between points. 
For convenience in the notation, $\mathbf{G}(\mathbf{Y},\mathbf{Y})$ is denoted as $\mathbf{G}$.
For an in-depth derivation of CPD, please refer to~\cite{myronenko2010point}.

\subsection{Shape Space}
\label{sec:shape_space}
\begin{figure*}
	\centering
	\includegraphics[width=\linewidth]{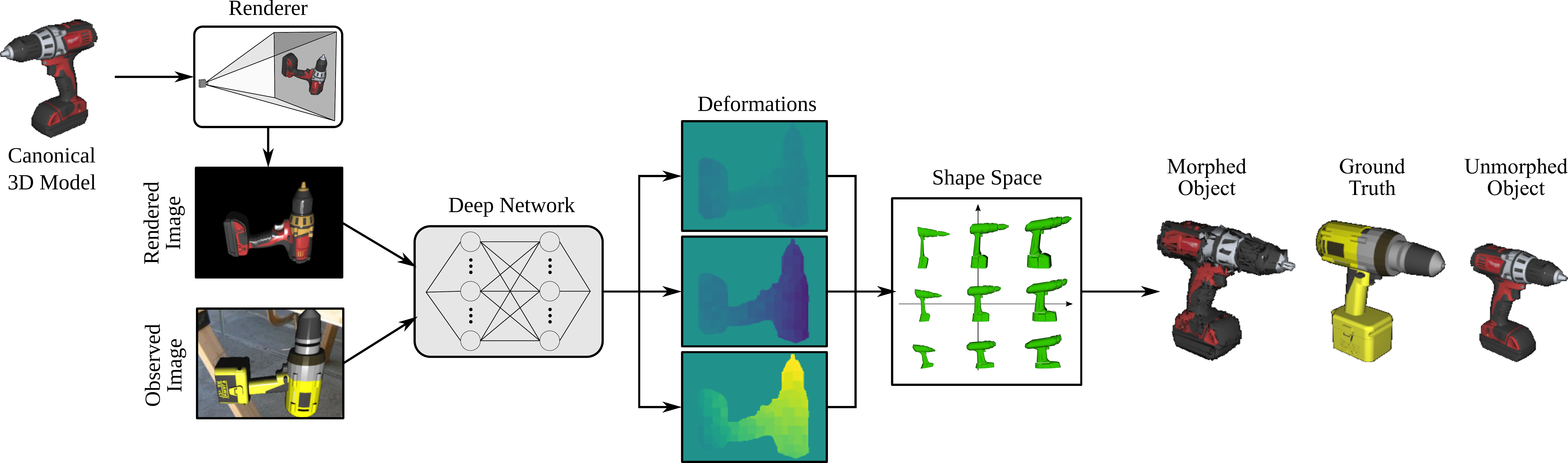}
	\caption{A canonical model is rendered in the same pose as the observed instance.
		Then, a deep CNN infers the $x$, $y$, and $z$ deformation components for each pixel of the rendered image.
		The deformations of the occluded parts are inferred by a shape space of the object category.
		Thanks to the shape space, the object is reconstructed (morphed object),
		which is shown from a non-observable viewpoint.
		For comparison, the ground truth object and the unmorphed canonical model are shown at the rightmost.}
	\label{fig:overview}
	\vspace*{-2ex}
\end{figure*}

For a known object category,
we construct a shape space that describes typical geometrical variations of the category.
An object category is defined as a set of 3D object models with similar extrinsic geometry.
In order to ensure consistency between the deformations, 
all the 3D models are aligned to a common coordinate frame.
Initially, 
a canonical model $\mathbf{C}$ is chosen and each instance $i$ is deformed against this canonical model following:
\begin{equation}
	\label{eq:trafos}
	\mathcal{T}_i(\mathbf{C}, \mathbf{W}_i) = \mathbf{C} + \mathbf{G}(\mathbf{C}, \mathbf{C}) \mathbf{W}_i.
\end{equation}
The shape space is defined as a low-dimensional space spanned by the principal components of all deformation fields $\mathbf{W}_i$.
This is possible based on the observation that the $\mathbf{G}$ matrix only depends on the point set that is deformed,
i.e.,
the canonical model.
In other words, 
the uniqueness of an object deformation inside a category is fully captured by its $\mathbf{W}_i$ matrix.

Note that the dimensionality of all $\mathbf{W}_i$ matrices is the same. 
This allows us to define feature vectors $\mathbf{w}_i \in \mathbb{R}^{3n}$,
and perform the Principal Component Analysis Expectation Maximization (PCA-EM) to find the latent space of the category.
As result, 
a matrix $\mathbf{L} \in \mathbb{R}^{l\times 3n}$ containing $l$ principle components is determined. 
Any point $\mathbf{x} \in \mathbb{R}^{l}$ in the latent space can now be mapped into a feature vector $\hat{\mathbf{w}} \in \mathbb{R}^{3n}$ by:
\begin{equation}
	\label{eq:wxL}
	\hat{\mathbf{w}} = \mathbf{L} \mathbf{x} + \bar{\mathbf{w}},
\end{equation}
where $\bar{\mathbf{w}}$ is the mean of all feature vectors. 
In this manner, the canonical matrix $\mathbf{C}$ together with the principle components $\mathbf{L}$ represent the deformation model for an object category. 
\section{Method}
\label{sec:method}
In this section, we describe our approach that deforms a canonical 3D model of an object category into a novel instance observed by a single RGB image (Fig.~\ref{fig:overview}).
Our approach is composed of three main components: a renderer~\cite{schwarz2020}, a deep CNN, and a shape space.
The renderer is in charge of generating 2D images of realistic 3D models which will be used to train the CNN.
Given the images of the canonical and observed models,
the CNN infers a three-channel image that represents deformation vectors for each of the visible pixels of the image of the canonical model.
The final deformation field is estimated by searching in the shape space for a feature vector whose deformation field matches best the deformations inferred by the network.
By means of the shape space, an estimation of the deformation field of the non-observable parts is inferred.

\subsection{Deformation Representation}
\label{sec:defor_rep}
For a given object category,
e.g., \textit{drill},
a canonical model is chosen.
This model consists of a textured three-dimensional mesh. 
Model point clouds can be generated by ray-casting from several viewpoints on a tessellated sphere followed by down-sampling, e.g., by a voxel grid filter.
The matrix of the point cloud of the canonical model is referred as $ \mathbf{C} \in \mathbb{R}^{n \times 3} $,
while the mesh point matrix is denoted $\mathbf{C}_m \in \mathbb{R}^{m \times 3} $. 
The point cloud matrix is used to define a deformation on the mesh. 
The matrix of the deformed mesh $\mathbf{C}_m^\prime \in \mathbb{R}^{m \times 3}$ is defined as:
\begin{equation}
	\label{eq:deformation}
	\mathbf{C}_m^\prime = \mathbf{C}_m + \mathbf{G}(\mathbf{C}_m, \mathbf{C}) \mathbf{W}(\mathbf{C}, \mathbf{O}),
\end{equation}
where $\mathbf{W}(\mathbf{C}, \mathbf{O}) \in \mathbb{R}^{n \times 3}$ describes the offsets that should be applied to the points of the canonical point cloud $\mathbf{C}$ to deform it towards the point cloud of the observed instance represented by $\mathbf{O} \in \mathbb{R}^{k \times 3}$. 
The offsets $\mathbf{W}(\mathbf{C}, \mathbf{O})$ are multiplied by $\mathbf{G}(\mathbf{C}_m, \mathbf{C}) \in \mathbb{R}^{m \times n}$ to map them into the offsets of the mesh vertices.
$\mathbf{G}(\mathbf{C}_m, \mathbf{C})$ is calculated as described in Eq. \eqref{eq:G} and ensures coherent movement of the vertices. 
Note that $\mathbf{W}(\mathbf{C}, \mathbf{O})$ is the only part of Eq.~\eqref{eq:deformation} that depends on the observed instance. 

To generate target images for training the CNN, 
we define the offset matrix $\boldsymbol{\delta}(\mathbf{C}, \mathbf{O})\in \mathbb{R}^{n\times 3}$:
\begin{equation}
	\label{eq:off_im}
	\boldsymbol{\delta}(\mathbf{C}, \mathbf{O}) = \mathbf{G}(\mathbf{C}, \mathbf{C}) \mathbf{W}(\mathbf{C}, \mathbf{O}).
\end{equation}
and represent it as a three channel image, 
where each channel describes the deformations in one of the coordinate axis: $x$, $y$ and $z$.
The mapping from the 3D offset matrix $\mathbf{\delta}$ to the image space is described in Sec.~\ref{sec:dataset}.

\subsection{Zoom Operation}
\label{sec:zoom}
\begin{figure}[tbh]
	\centering
	\includegraphics[width=\linewidth]{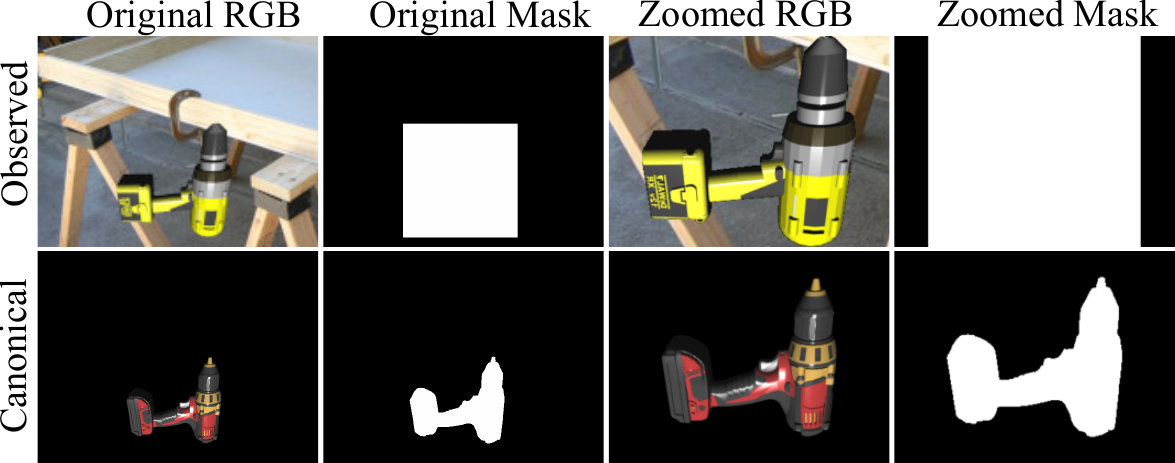}
	\caption{Zoom operation. 
		The canonical model is rendered at the same pose as the observed one.
		Both images are cropped according to a bounding box that contains both objects and has the aspect ratio of the network input.
		Finally, the cropped images are bilinear upsampled.}
	\label{fig:zoom}
	\vspace*{-2ex}
\end{figure}

Inspired by \cite{deepIm}, 
the amount of object details is increased by zooming in the observed and rendered images before feeding them into the network.
Given the pose of the observed object, 
the canonical model is rendered placing the object at the same place as the observed one with respect to the camera.
Then, a bounding box that contains both objects and has the same aspect ratio as the input images of the network is defined.
Following this bounding box, 
both images are cropped and upsampled bilinearly to the fixed size of the network input (256$\times$192 in our experiments).
An example of the zoom operation is shown in Figure~\ref{fig:zoom}.

\subsection{Convolutional Neural Network}
\label{sec:network}

The backbone of our network is the FlowNet2 architecture~\citep{flownet2} because we assume a connection between the optical flow among two objects and the point offsets that should be applied for a matching deformation. 
We modify FlowNet2 such that it receives two RGB and two mask images as input, 
and it returns a three channel image representing the deformation for each pixel.
The mask image of the canonical model is directly given by the renderer and serves as a regularizer to distinguish the foreground and background pixels
while the mask of the observed image is a bounding box containing the object. 
The mask images can be interpreted as a fourth channel additional to the RGB channels of the two input images.
Thus, we use the terms observed and canonical images in the remaining sections of this paper to refer to the network input.

\subsection{Deformation Inference of Occluded Parts}
\label{sec:occluded_parts}

Given a single RGB image, 
an object cannot be fully observed.
Thus, we first find deformations on the visible object pixels and use these deformations to infer the missing ones of the occluded parts.
Estimating deformations on the observable parts is handled by the network (Sec.\ref{sec:network}). 
For inferring deformations of the occluded parts, 
a shape space is constructed as explained in Sec.~\ref{sec:shape_space}. 
Then, we search for a feature vector $\hat{\mathbf{w}}$ in the shape (latent) space 
that matches best the deformations of the observable points. 

The CNN outputs a three channel image containing a deformation vector per pixel.
Now,
a mapping from the image space to the vector space of $\mathbf{C}$ needs to be established.
Thus,
we find the closest point on $\mathbf{C}$ to the position of each pixel in the position tensor given by the renderer.
In general, the number of pixels is larger than the number of points of $\mathbf{C}$, 
and multiple pixels are assigned to the same point of $\mathbf{C}$. 
Consequently, the deformation of the closest points of $\mathbf{C}$ is defined as the mean of the assigned deformations.
This results in a matrix that describes how the closest points of the canonical model have to be moved to match the shape of the visible parts of the observed object. 
For the occluded points, we assume an initial offset of zero and build a sparse matrix $\boldsymbol{\delta}_{vis} \in \mathbb{R}^{n\times 3}$ with the deformation of the closest points. 

\begin{figure}[]
	\centering
	\includegraphics[width=\linewidth]{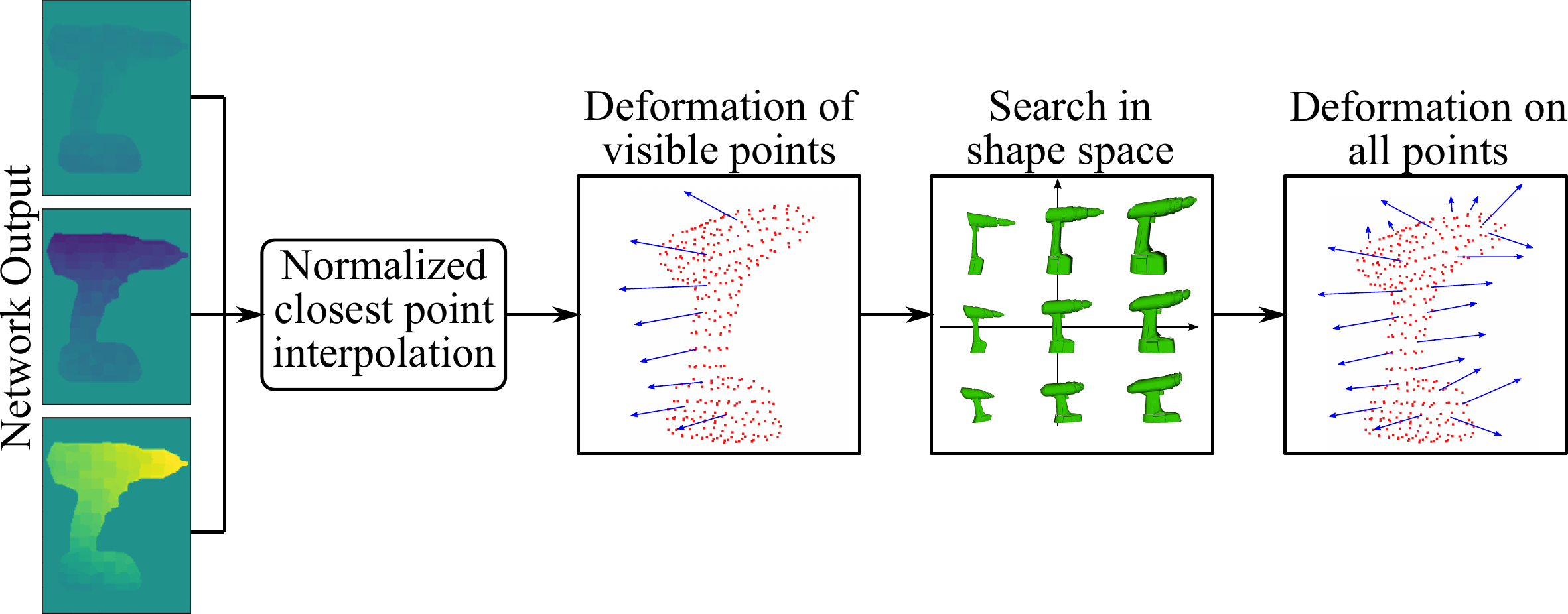}
	\caption{The network output defines a deformation field for a 3D point set generated by a normalized closest point interpolation.
		This point set represents the visible points of the canonical model.
		Deformation vectors for all points of $\mathbf{C}$ are inferred by means of the shape space of the category.}
	\label{fig:partial_to_full}
	\vspace*{-2ex}
\end{figure}

Obtaining a deformation vector for all points of $\mathbf{C}$ can be formulated as the task of searching for a feature vector $\mathbf{x}$ in the shape space that matches the inferred deformations of the visible points.
This feature vector $\mathbf{x}$ defines a deformation matrix:
\begin{equation}
	\hat{\boldsymbol{\delta}} = \tilde{\mathbf{G}}(\mathbf{C}, \mathbf{C}) (\mathbf{L} \mathbf{x} + \bar{\mathbf{w}}),
\end{equation}
where $\tilde{\mathbf{G}}(\mathbf{C}, \mathbf{C}) \in \mathbb{R}^{3n\times 3n}$ is a rearranged form of $\mathbf{G}(\mathbf{C}, \mathbf{C}) \in \mathbb{R}^{n\times n}$ with additional zeros to match the dimensionality of $\mathbf{L} \mathbf{x} + \bar{\mathbf{w}}\in \mathbb{R}^{3n\times 1}$.
Our objective is then to minimize the loss on the visible points between $\boldsymbol{\delta}_{vis}$ and $\boldsymbol{\hat{\delta}}$:
\begin{equation}
	\label{eq:loss1}
	L(\boldsymbol{\delta}_{vis}, \boldsymbol{\hat{\delta}}) = \sum_{v \in i_{vis}} \left\lVert \boldsymbol{\delta}_{vis}(v,.) - \boldsymbol{\hat{\delta}}(v,.)  \right\rVert ^ 2,
\end{equation}
where $i_{vis}$ describes the set of indices belonging to visible points and $\boldsymbol{\delta}(v,.)$ describes the $v$-th row of matrix $\boldsymbol{\delta}$. 
This minimization problem can be expressed as a least squares problem. 
Let $n_v$ denote the number of visible points
and let define $\overline{\boldsymbol{\delta}}_{vis} \in \mathbb{R}^{3n_v}$ as $\boldsymbol{\delta}_{vis}$ after removing every row containing zeros (occluded points) and rearranging it as a row vector. 
We introduce a matrix $\mathbf{D} \in \mathbb{R}^{n_v \times n}$ which removes exactly the same rows from $\boldsymbol{\hat{\delta}}$ as the ones removed from $\boldsymbol{\delta}_{vis}$. 
In this manner, 
minimizing Eq.~\eqref{eq:loss1} results in:
\begin{equation}
	L(\overline{\boldsymbol{\delta}}_{vis}, \mathbf{x}) =\left\lVert \overline{\boldsymbol{\delta}}_{vis} - \mathbf{D}(\tilde{\mathbf{G}}(\mathbf{C}, \mathbf{C}) ( \mathbf{L} \mathbf{x} + \bar{\mathbf{w}} ) ) \right\rVert ^ 2.
\end{equation}
Defining:
\begin{equation} 
\mathbf{A} := (\mathbf{D}\tilde{\mathbf{G}}(\mathbf{C}, \mathbf{C}) \mathbf{L}),
\end{equation}
and
\begin{equation}  
\mathbf{B} := (\overline{\boldsymbol{\delta}}_{vis} - \mathbf{D}\tilde{\mathbf{G}}(\mathbf{C}, \mathbf{C}) \bar{\mathbf{w}}),
\end{equation}
the feature vector $\mathbf{x}^{*}$ that represents the deformation of all points in $\mathbf{C}$ is found by:
\begin{equation}
\label{eq:final}
\mathbf{x}^{*}=\argmin_{\mathbf{x}} {L(\overline{\boldsymbol{\delta}}_{vis}, \mathbf{x}) = \argmin_{\mathbf{x}} \left\lVert \mathbf{A} \mathbf{x} - \mathbf{B }\right\rVert ^ 2}.
\end{equation}
Eq.~\eqref{eq:final} can now be solved by an off-the-shelf linear solver.
By using Eq.~\eqref{eq:wxL} with $\mathbf{x}^{*}$ and the corresponding rearrangement,
the final deformation field $\hat{\mathbf{W}}^{*}$ is obtained.
Moreover, by using Eq.~\eqref{eq:deformation}, 
the observed model is reconstructed.
A schematic overview of the deformation inference of occluded parts can be seen in Figure \ref{fig:partial_to_full}.
\begin{figure}[b!]
	\vspace*{-2ex}
	\centering
	\includegraphics[width=\linewidth]{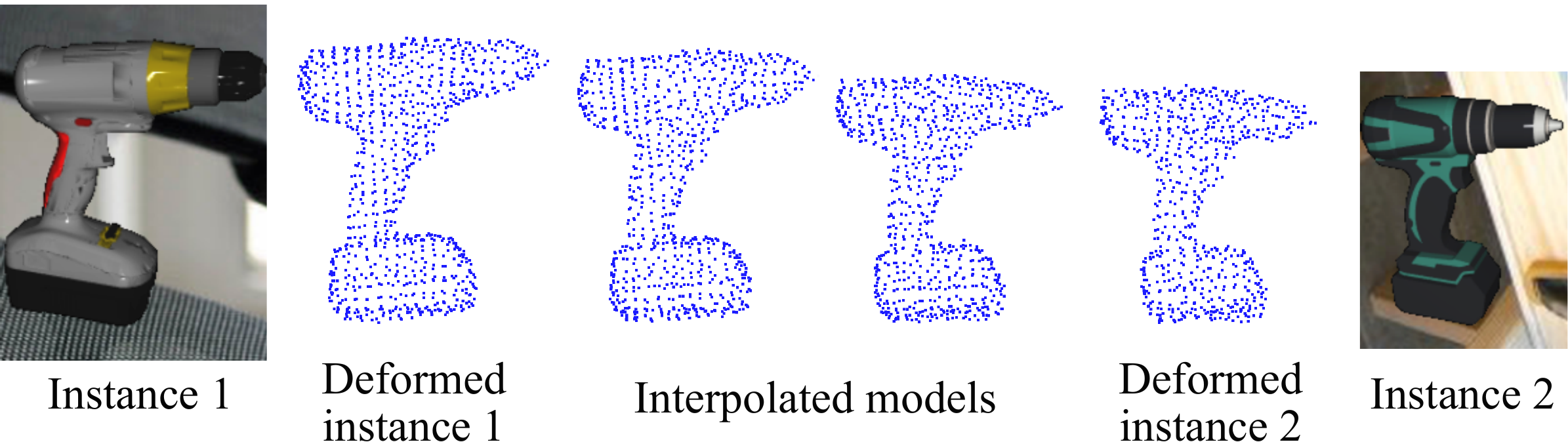}
	\caption{Registration of two observed instances against each other. 
		The resulting deformed models are interpolated for visualization.
	}
	\label{fig:reg_two_instances}
\end{figure}

\begin{figure*}[tbh]
	\centering
	\includegraphics[height=0.09\linewidth, angle=90, trim={300pt 170 30pt 120}, clip]{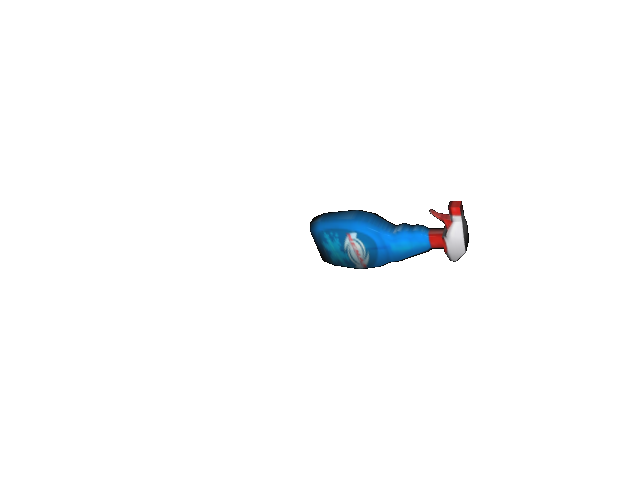}
	\includegraphics[height=0.09\linewidth, angle=90, trim={300pt 170 30pt 120}, clip]{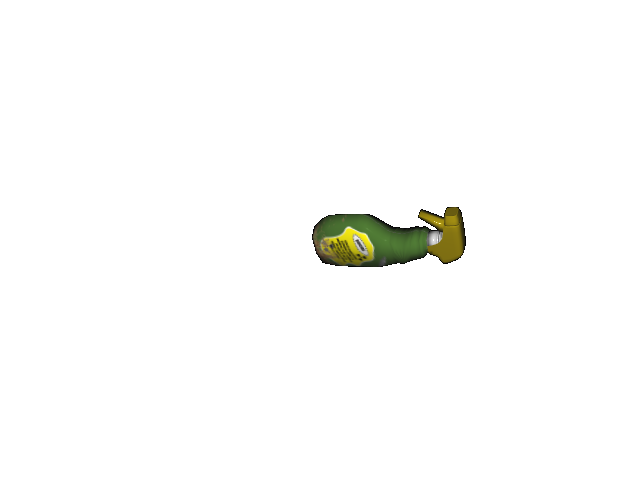}
	\includegraphics[height=0.09\linewidth, angle=90, trim={300pt 170 30pt 120}, clip]{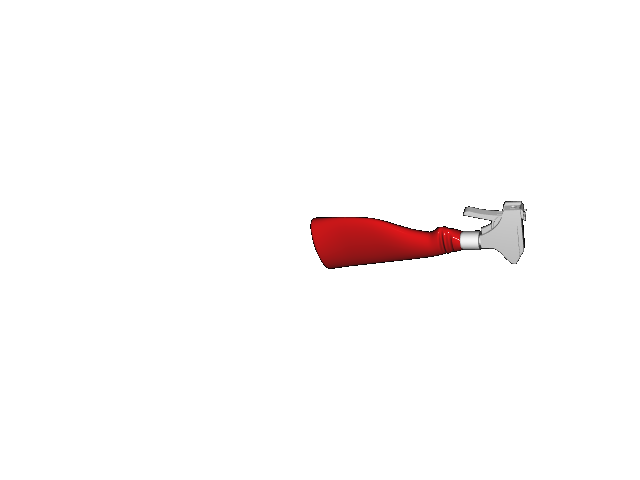}
	\includegraphics[height=0.09\linewidth, angle=90, trim={300pt 170 30pt 120}, clip]{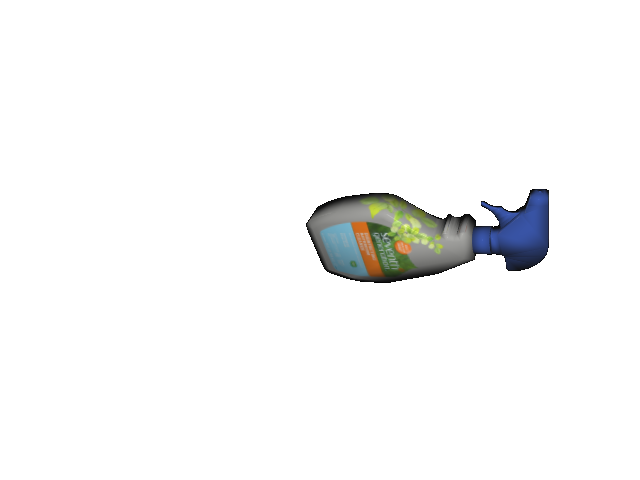}
	\includegraphics[height=0.09\linewidth, angle=90, trim={300pt 170 30pt 120}, clip]{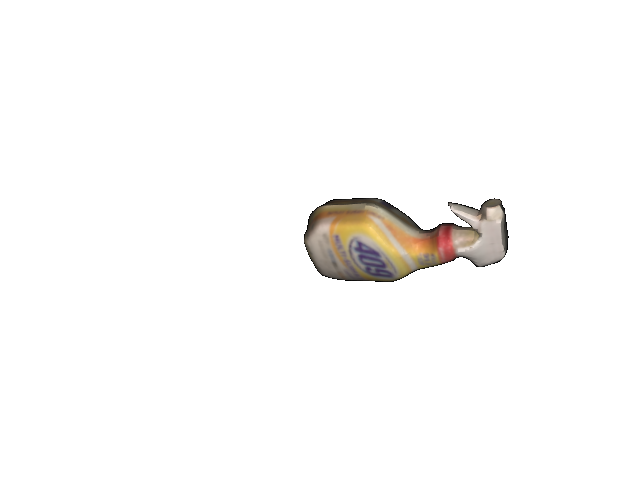}
	\includegraphics[height=0.09\linewidth, angle=90, trim={300pt 170 30pt 120}, clip]{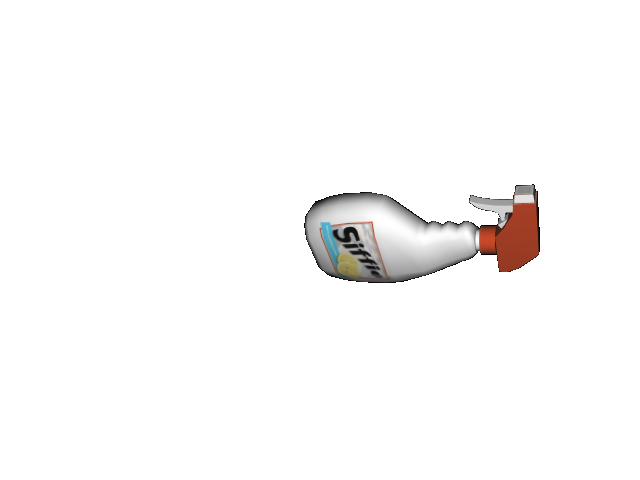}
	\includegraphics[height=0.09\linewidth, angle=90, trim={300pt 170 30pt 120}, clip]{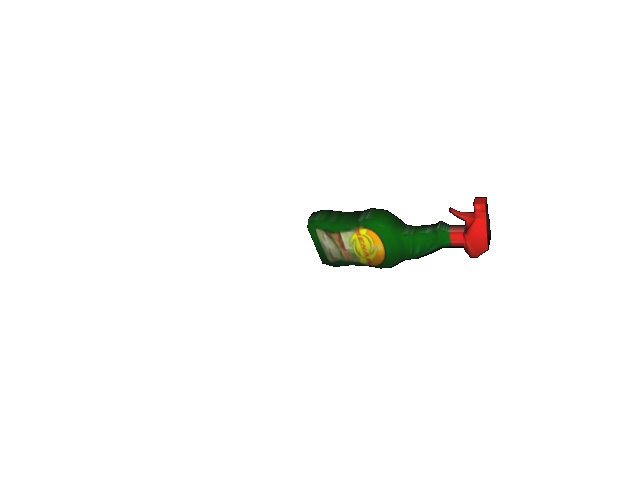}
	\includegraphics[height=0.09\linewidth, angle=90, trim={300pt 170 30pt 120}, clip]{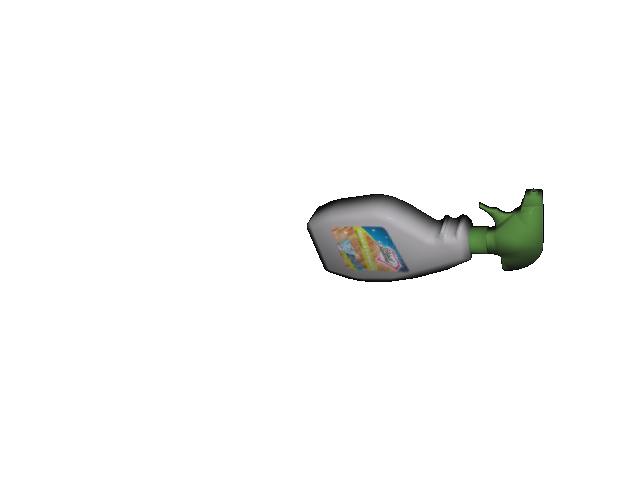}
	\includegraphics[height=0.09\linewidth, angle=90, trim={300pt 170 30pt 120}, clip]{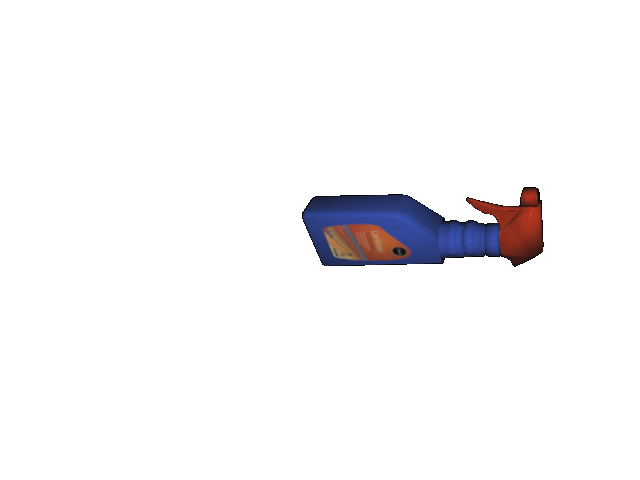}
	\includegraphics[height=0.09\linewidth, angle=90, trim={300pt 170 30pt 120}, clip]{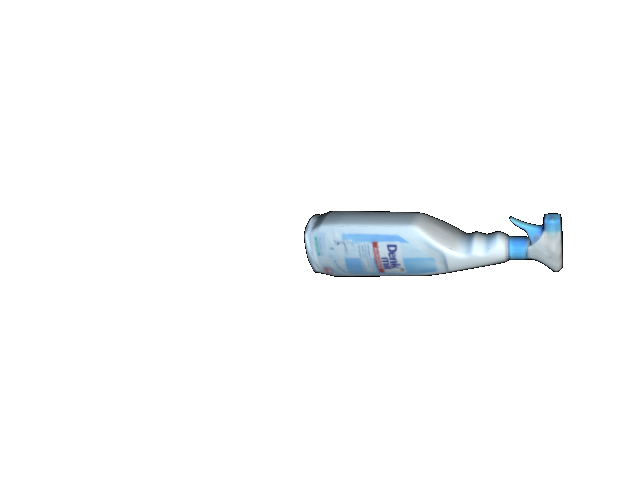}	
	
	\includegraphics[height=0.09\linewidth, angle=90, trim={240pt 150 190pt 100}, clip]{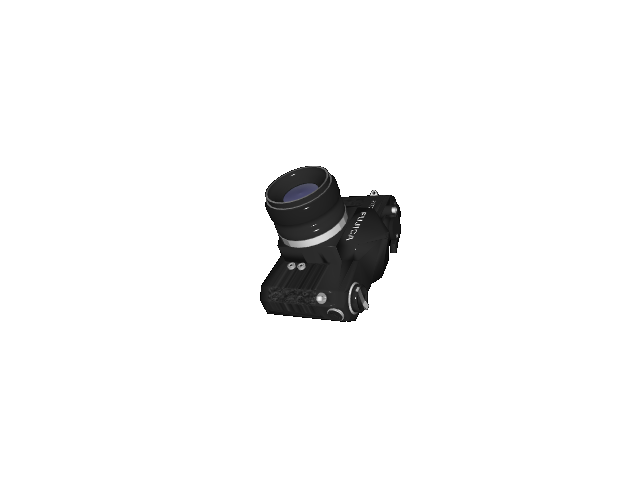}
	\includegraphics[height=0.09\linewidth, angle=90, trim={240pt 150 190pt 100}, clip]{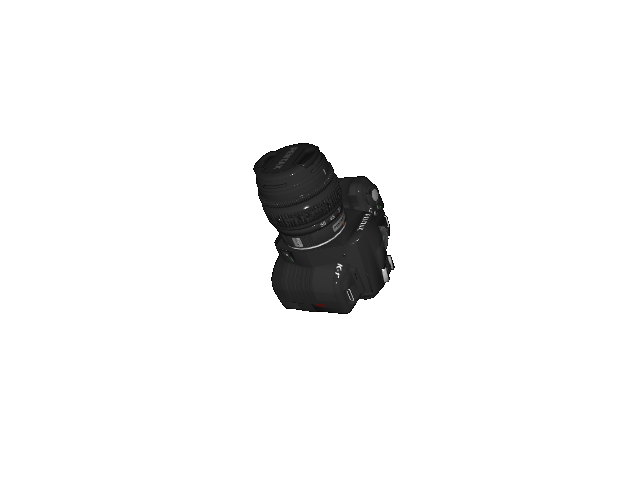}
	\includegraphics[height=0.09\linewidth, angle=90, trim={240pt 150 190pt 100}, clip]{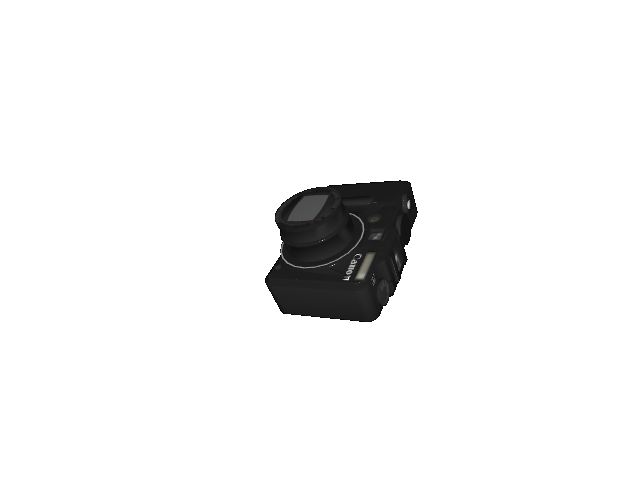}
	\includegraphics[height=0.09\linewidth, angle=90, trim={240pt 150 190pt 100}, clip]{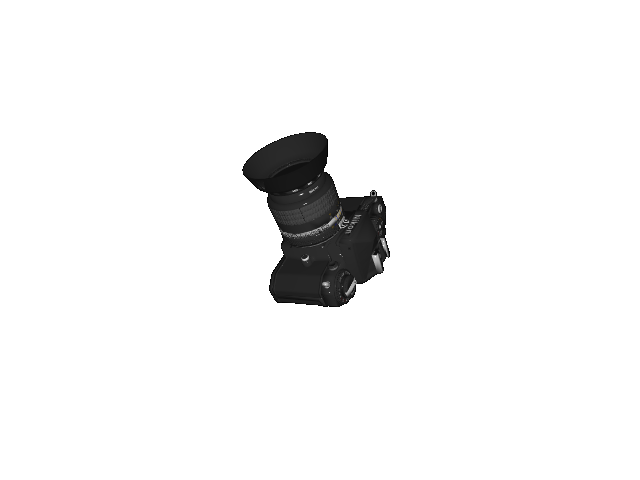}
	\includegraphics[height=0.09\linewidth, angle=90, trim={240pt 150 190pt 100}, clip]{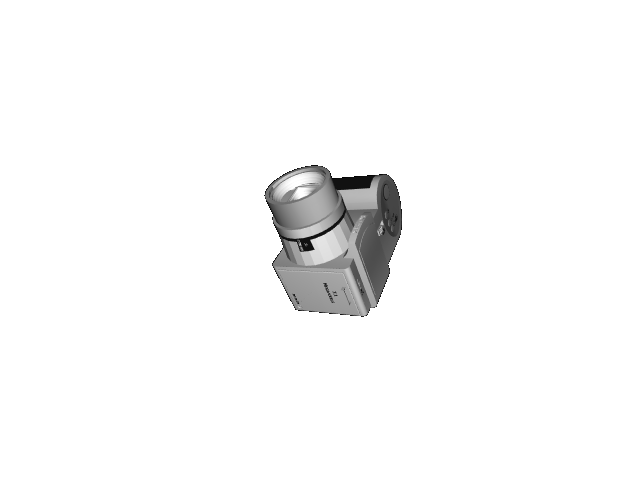}
	\includegraphics[height=0.09\linewidth, angle=90, trim={240pt 150 190pt 100}, clip]{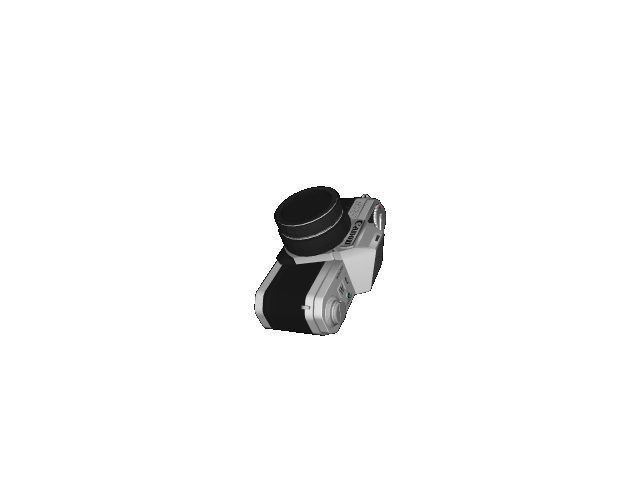}
	\includegraphics[height=0.09\linewidth, angle=90, trim={240pt 150 190pt 100}, clip]{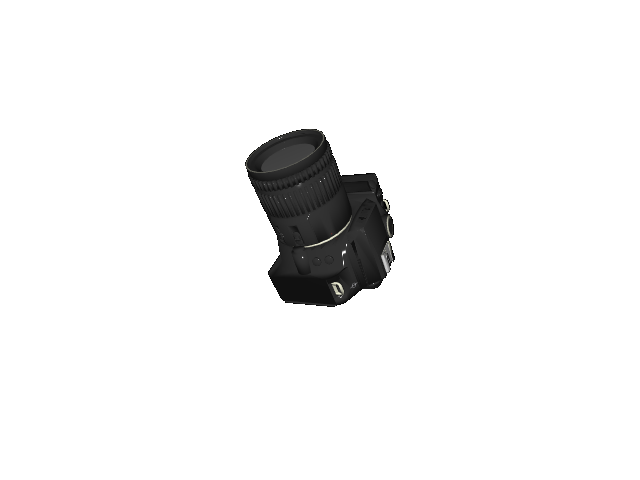}
	\includegraphics[height=0.09\linewidth, angle=90, trim={240pt 150 190pt 100}, clip]{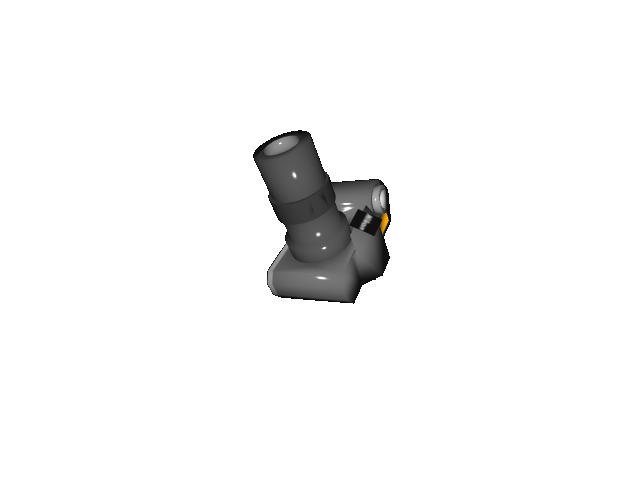}
	\includegraphics[height=0.09\linewidth, angle=90, trim={240pt 150 190pt 100}, clip]{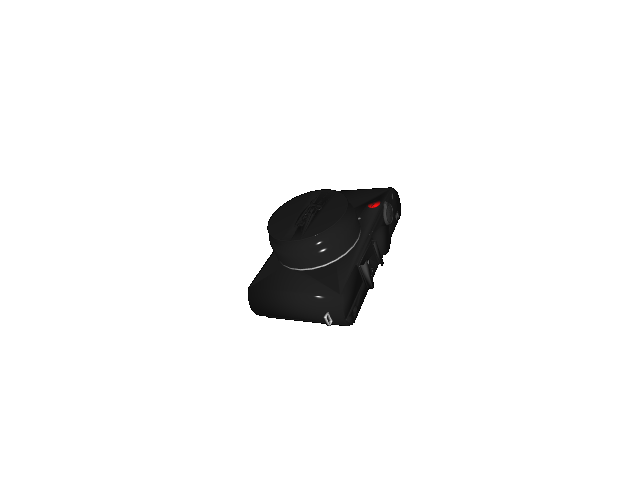}
	\includegraphics[height=0.09\linewidth, angle=90, trim={240pt 150 190pt 100}, clip]{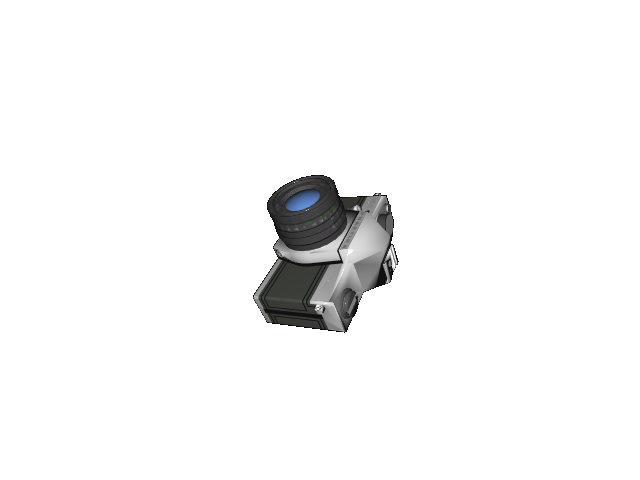}	
	
	\includegraphics[height=0.09\linewidth, angle=90, trim={300pt 170 100pt 100}, clip]{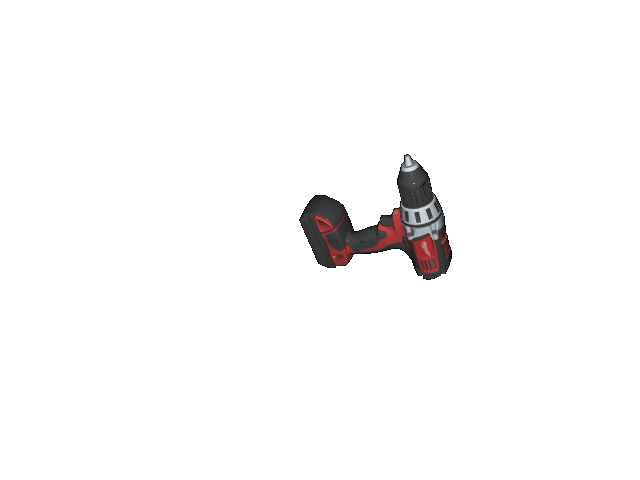}
	\includegraphics[height=0.09\linewidth, angle=90, trim={300pt 170 100pt 100}, clip]{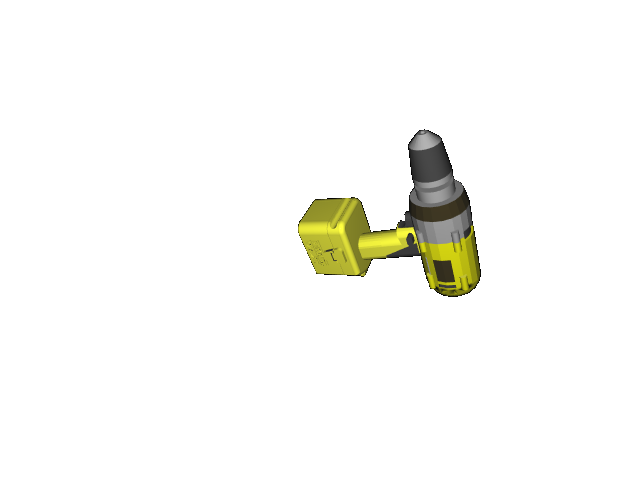}
	\includegraphics[height=0.09\linewidth, angle=90, trim={300pt 170 100pt 100}, clip]{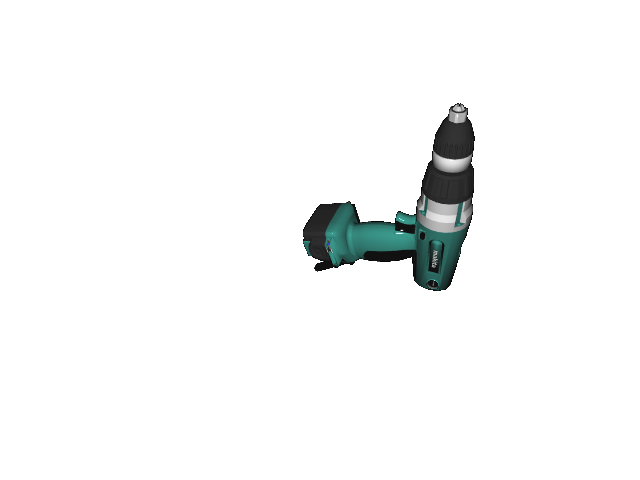}
	\includegraphics[height=0.09\linewidth, angle=90, trim={300pt 170 100pt 100}, clip]{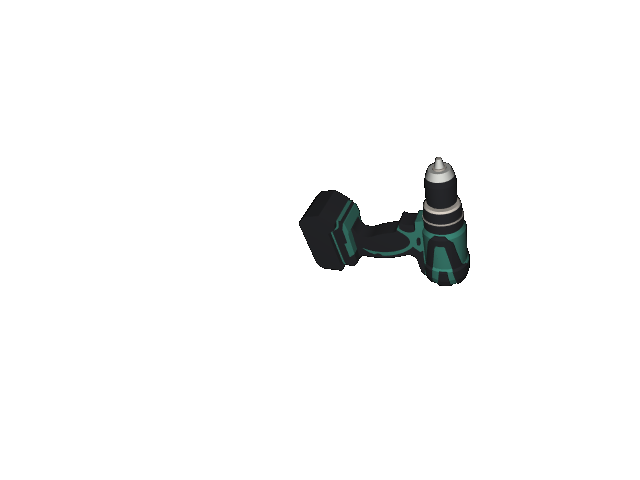}
	\includegraphics[height=0.09\linewidth, angle=90, trim={300pt 170 100pt 100}, clip]{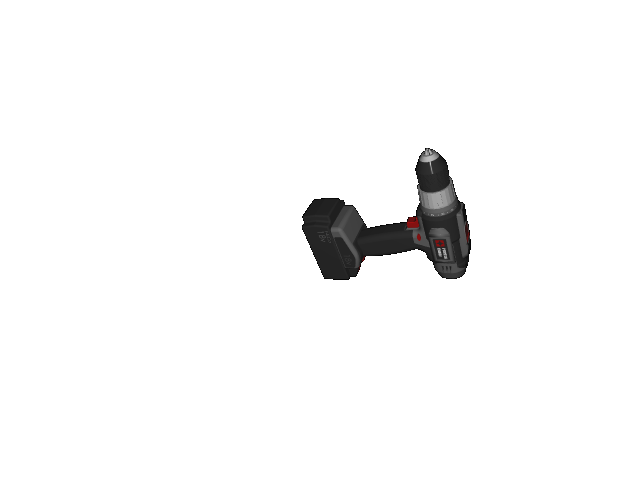}
	\includegraphics[height=0.09\linewidth, angle=90, trim={300pt 170 100pt 100}, clip]{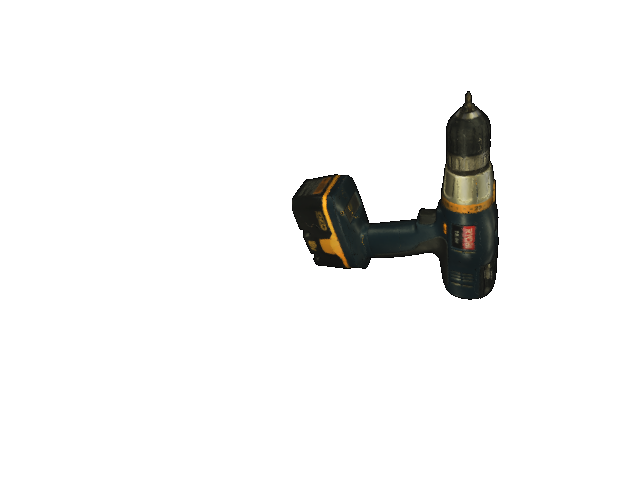}
	\includegraphics[height=0.09\linewidth, angle=90, trim={300pt 170 100pt 100}, clip]{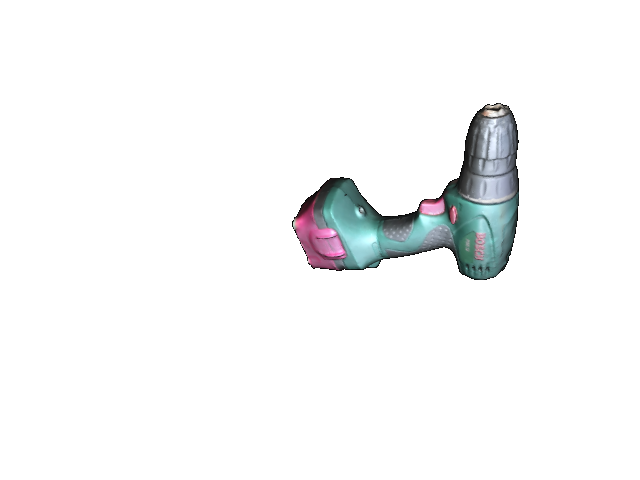}
	\includegraphics[height=0.09\linewidth, angle=90, trim={300pt 170 100pt 100}, clip]{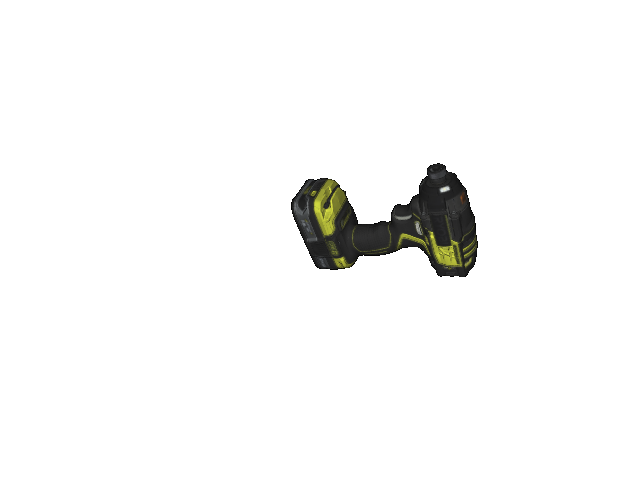}
	\includegraphics[height=0.09\linewidth, angle=90, trim={300pt 170 100pt 100}, clip]{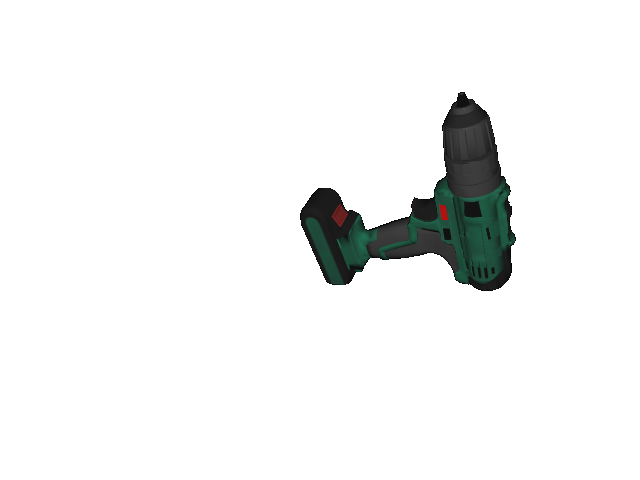}
	\includegraphics[height=0.09\linewidth, angle=90, trim={300pt 170 100pt 100}, clip]{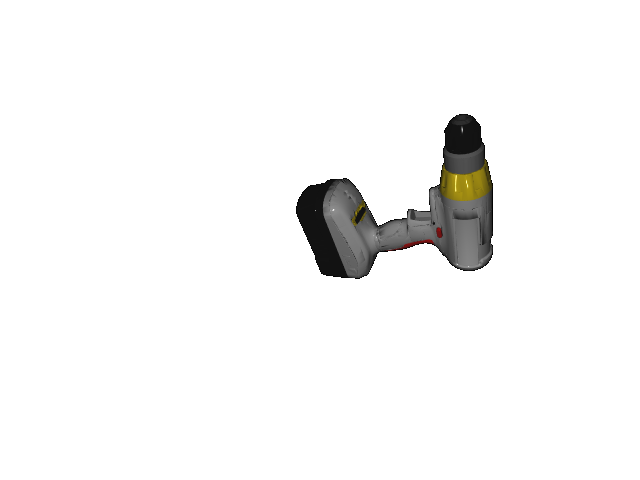}	
	
	\includegraphics[height=0.09\linewidth, angle=90, trim={160pt 90pt 60pt 90pt}, clip]{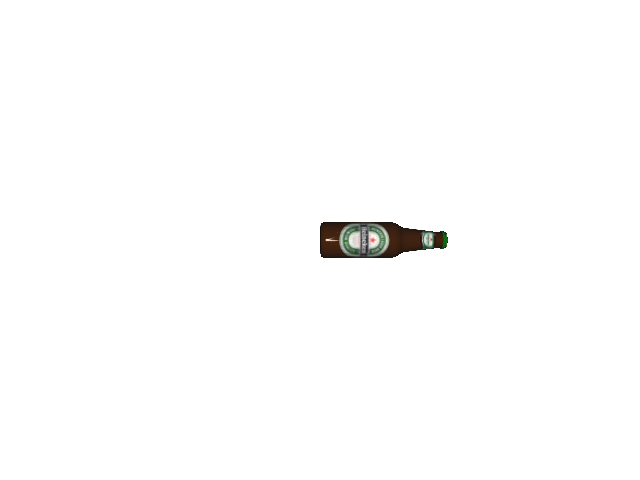}
	\includegraphics[height=0.09\linewidth, angle=90, trim={160pt 90pt 60pt 90pt}, clip]{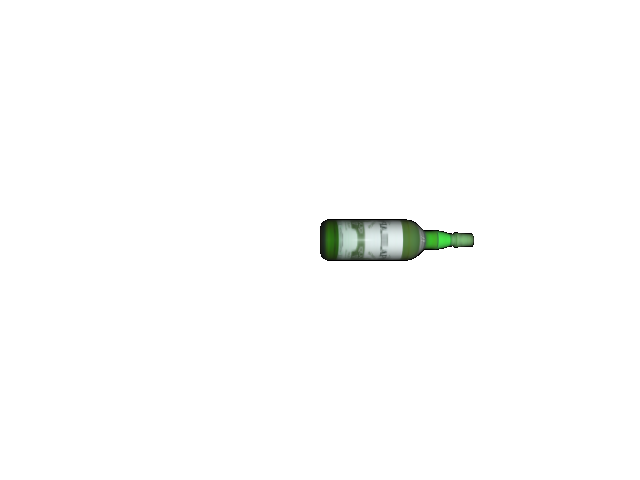}
	\includegraphics[height=0.09\linewidth, angle=90, trim={160pt 90pt 60pt 90pt}, clip]{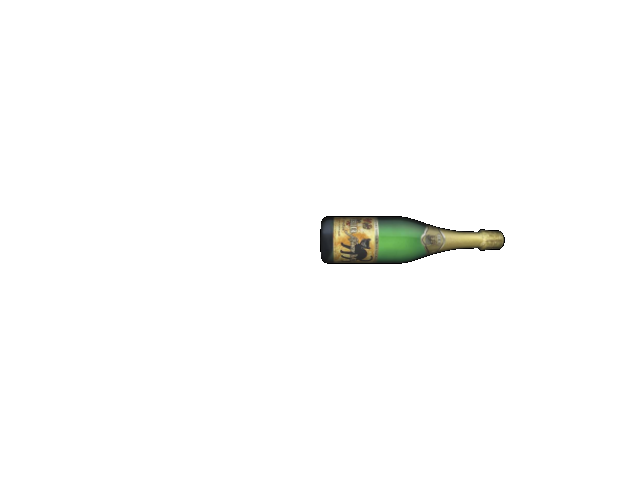}
	\includegraphics[height=0.09\linewidth, angle=90, trim={160pt 90pt 60pt 90pt}, clip]{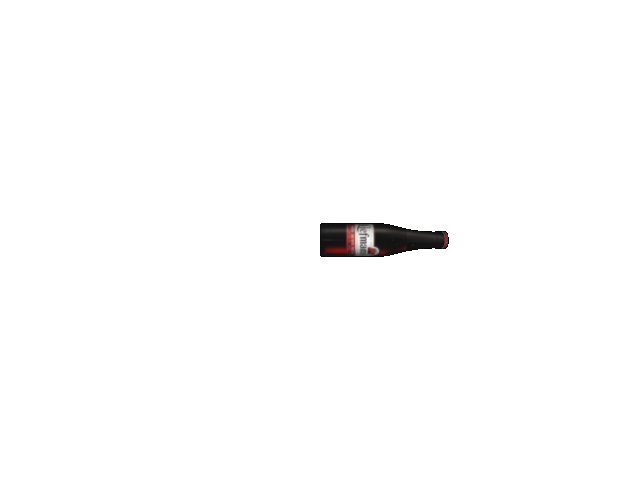}
	\includegraphics[height=0.09\linewidth, angle=90, trim={160pt 90pt 60pt 90pt}, clip]{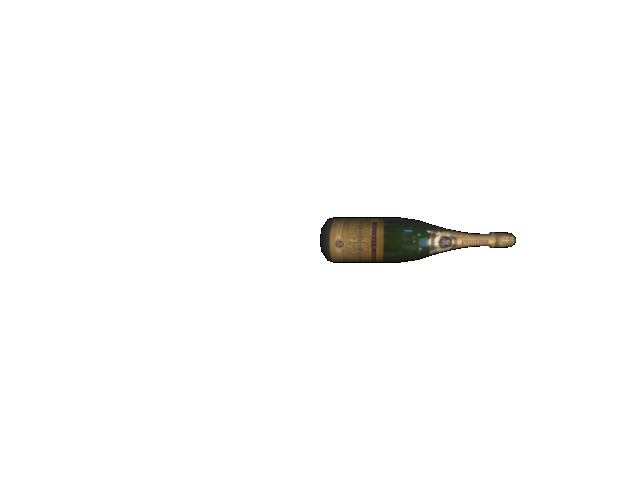}
	\includegraphics[height=0.09\linewidth, angle=90, trim={160pt 90pt 60pt 90pt}, clip]{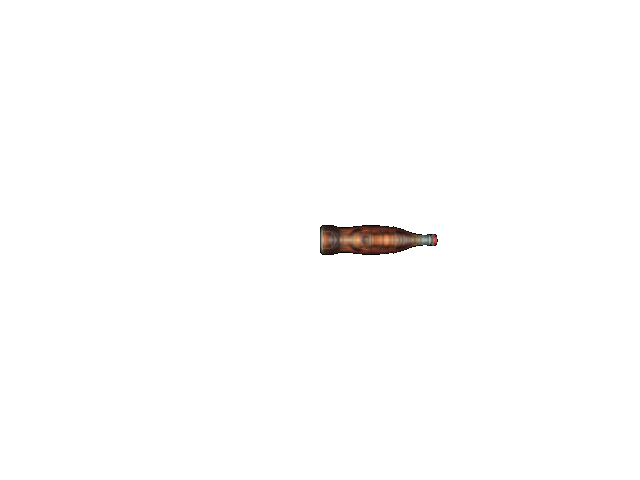}
	\includegraphics[height=0.09\linewidth, angle=90, trim={160pt 90pt 60pt 90pt}, clip]{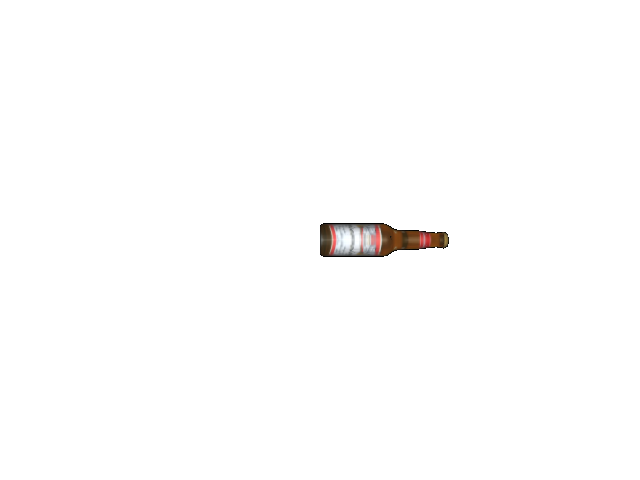}
	\includegraphics[height=0.09\linewidth, angle=90, trim={160pt 90pt 60pt 90pt}, clip]{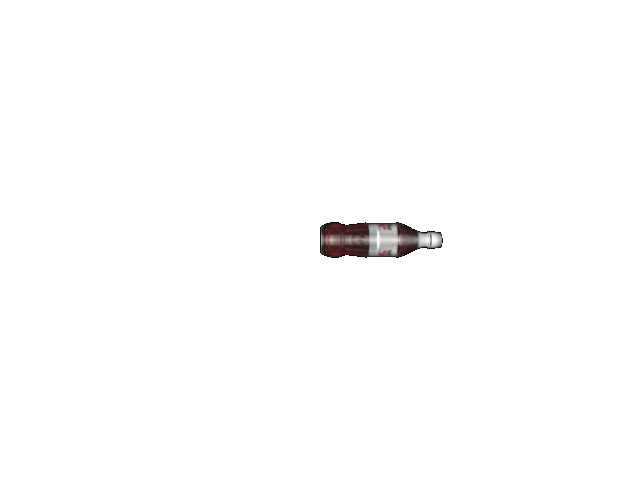}
	\includegraphics[height=0.09\linewidth, angle=90, trim={160pt 90pt 60pt 90pt}, clip]{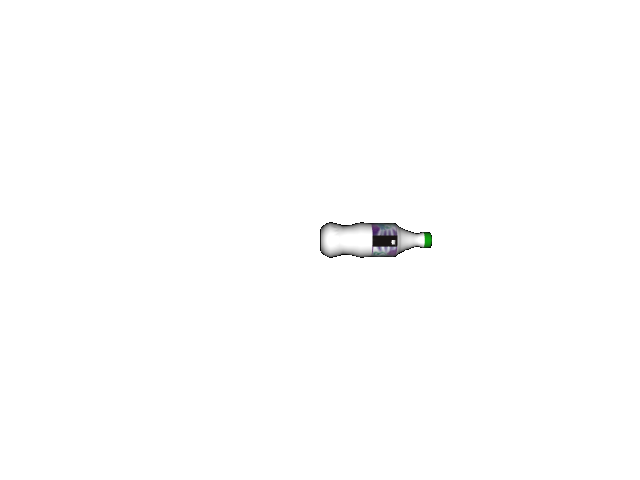}
	\includegraphics[height=0.09\linewidth, angle=90, trim={160pt 90pt 60pt 90pt}, clip]{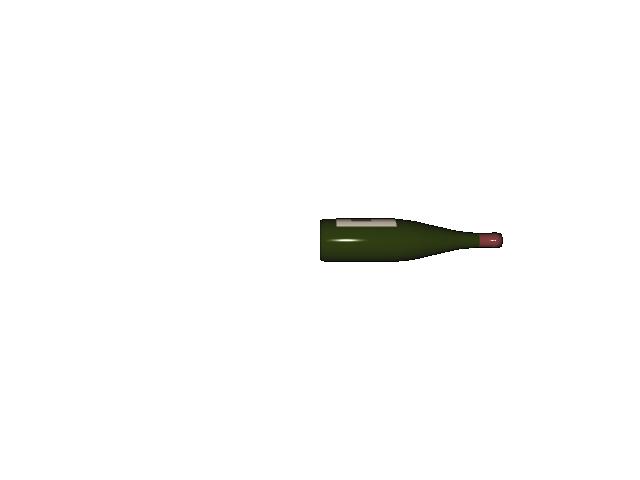}
	\caption{Samples of 3D models of the \textit{spray bottle}, \textit{camera} and \textit{drill}, and \textit{bottles} categories used in our experiments.
		The canonical models are leftmost.}
	\label{fig:models}
	\vspace*{-1ex}
\end{figure*}

Note that two observed instances can be registered against each other by simply finding their respective deformation fields towards the canonical model, 
since the dimensionality of the deformed objects is defined only by the shared canonical model (Fig.~\ref{fig:reg_two_instances}). 
\section{Dataset}
\label{sec:dataset}
To the best of our knowledge, 
there does not exist an available dataset that contains intra-class object deformations suitable for training CNNs.
In this section,
we explain how the dataset is generated.
Our dataset is composed of a set of 3D textured meshes that produce a set of rendered images and ground truth deformations for each object category.

To build the dataset, we collected realistic textured 3D meshes of different objects from the same object category.
The models were taken from online databases\footnote{\url{https://sketchfab.com}} and from~\cite{Wang_2019_CVPR}.
A canonical model is selected and separated from the other $k$ models. 
This canonical model and the collection of 3D data is also used for building the latent space for inferring deformations of the occluded parts (Sec.~\ref{sec:occluded_parts}).
Ground truth deformations from the canonical model towards all the other models are calculated using CPD (Sec.~\ref{sec:cpd}).
The objects are observed from different viewpoints to generate the rendered images.

\begin{figure}[b!]
	\centering
	\includegraphics[width=\linewidth]{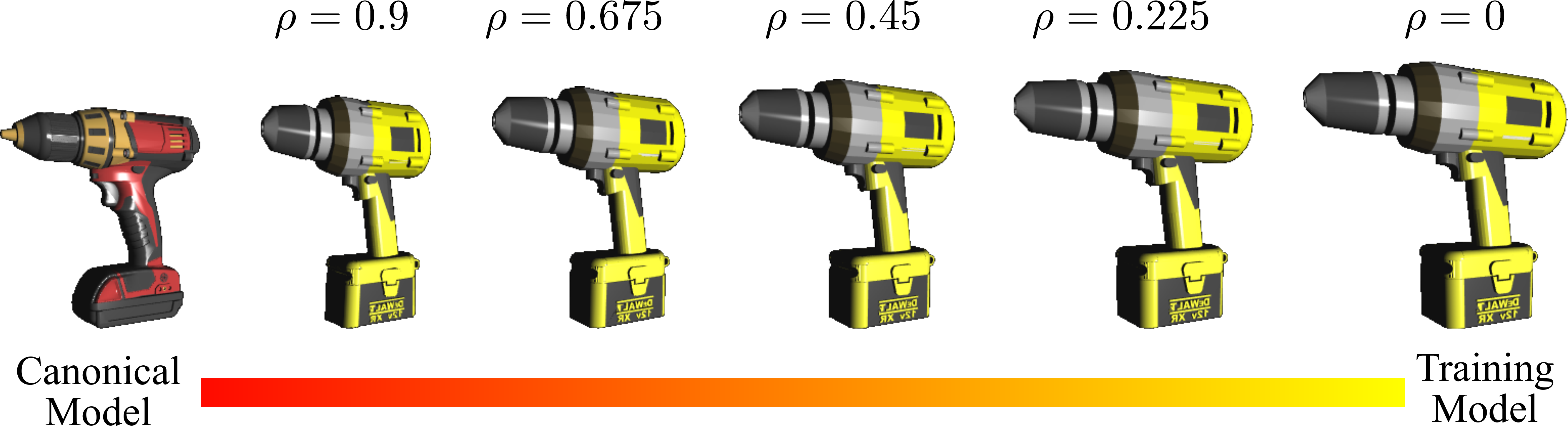}
	\caption{Interpolation process from the observed instance on the right towards the canonical instance on the left.}
	\label{fig:interpolate}
	\vspace*{-1ex}
\end{figure}

Since good quality 3D textured models are scarce,
specially graspable objects,
we interpolate models that are in between the canonical and the observed models (Fig.~\ref{fig:interpolate}).
To achieve this, 
$\mathcal{T}(\mathbf{C}, \mathbf{W}_i)$ is calculated for all the testing models, as described in Eq.~\eqref{eq:trafos}. 
Note that simply using Eq.~\eqref{eq:deformation} to interpolate from the canonical and to the observed models will generate instances with the same texture as the canonical mesh,
reducing the generalization capabilities of the network.
Therefore, 
an inverse deformation (from observed to canonical) is defined based on the point cloud of the observed training instance $\mathbf{T}_i$ and the deformations fields $\mathbf{W}_i$:
\begin{equation}
	\label{eq:int}
	\mathcal{T}^{-1}(\mathbf{T}_i, \mathbf{W}_i) = \mathbf{T}_i + \mathbf{G}(\mathbf{T}_i, \mathbf{C}) (-\mathbf{W}_i).
\end{equation}
Similarly to Eq.~\eqref{eq:deformation}, 
the vertices $\mathbf{T}_{m,i}$ of the observed training instances are morphed following:
\begin{equation}
	\mathbf{T}_{m,i}^\prime = \mathbf{T}_{m,i} + \mathbf{G}(\mathbf{T}_{m,i}, \mathbf{C}) (-\mathbf{W}_i).
\end{equation}
Finally, by adding an interpolation factor $\rho$ we obtain:
\begin{equation}
	\label{eq:morph}
	\mathbf{T}_{m,i}^\prime(\rho) = \mathbf{T}_{m,i} + \mathbf{G}(\mathbf{T}_{m,i}, \mathbf{C}) (-\rho\mathbf{W}_i).
\end{equation}
Eq.~\eqref{eq:morph} allows us to interpolate between every observed model and the canonical model without incurring in expensive computations of CPD for each interpolation model.

\begin{figure*}
	\centering
	\includegraphics[width=0.85\linewidth]{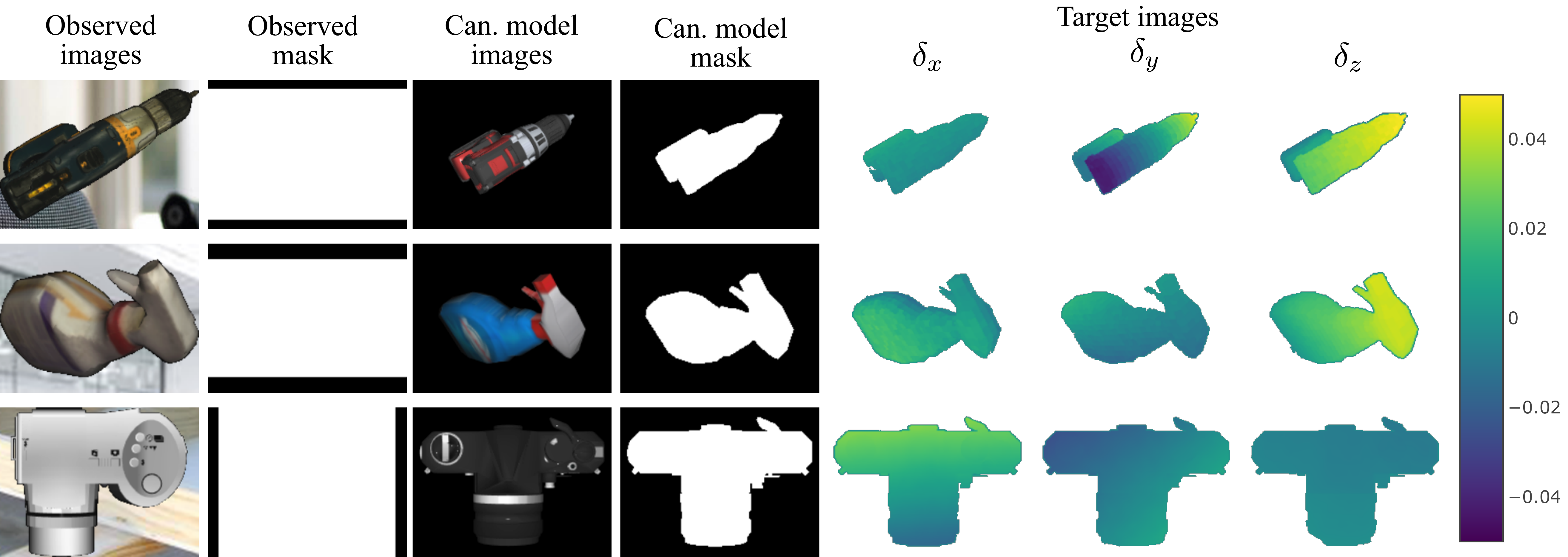}
	\caption{Generated training images. 
		The networks receive four images: an observed RGB image with its bounding box mask image,
		and a rendered image of the canonical model with its mask image.
		The target is a three channel image containing the $x$, $y$ and $z$ component of the deformations which are represented using heatmaps.
		For clarity in the visualization, the background of the target images are painted white.}
	\label{fig:generated_imgs}
	\vspace*{-1ex}
\end{figure*}
For a given observed model $i$ and a given interpolation constant $\rho$, 
we morph the model according to Eq.~\eqref{eq:morph}.
This morphed model and the canonical model are rendered from different viewpoints.
We generate a bounding box for the image of the observed model and a mask for the rendered canonical image. 
This four images are then zoomed in as explained in Sec.~\ref{sec:zoom}.

Apart from the RGB image,
the renderer also provides a three channel position image.
This image contains in each pixel the information about the $x$, $y$, and $z$ coordinates of the visible parts of the rendered model.
The position image is cropped and upsampled in the same way as the other four images.
The resulting position tensor defines a 3D point set $\mathbf{P}$.
Given the canonical point set and the corresponding deformation vector for each point,
we perform a radial basis function interpolation with a linear kernel to find the deformation vector for each point of $\mathbf{P}$.
The results of the interpolation are then expressed as a three channel image,
each channel containing the $x$, $y$ and $z$ deformations for each pixel.
The ground truth target data for the network is generated as explained in Sec.~\ref{sec:defor_rep}.
However, because of the interpolated models, 
Eq.~\ref{eq:off_im} is expanded as:
\begin{equation}
	\boldsymbol{\delta}(\mathbf{C}, \mathbf{T}_i) = \mathbf{G}(\mathbf{C}, \mathbf{C}) (1-\rho) \mathbf{W}_i.
\end{equation}

\section{Evaluation}
\label{sec:evaluation}

\subsection{Experimental Setup}
\label{sec:experimental_setup}
We tested our approach on four categories: \textit{spray bottles}, \textit{cameras}, \textit{bottles}, and \textit{drills} containing 12, 15, 14 and 16 instances, respectively.
Figure~\ref{fig:models} shows 3D model samples of the categories used in the experiments.
For the dataset generation, we used four $\rho\in\{0.0, 0.25, 0.5, 0.75\}$ interpolation values and 74 viewpoints on a tessellated sphere.
Pulling apart one instance as the canonical model and two instances for testing results in 2664, 3552, 3256 and 3848 training images, respectively.
To obtain ground truth deformations we used CPD with $\lambda=2.0$ and $\beta=2.0$ across all object instances. 
All the shape spaces have $l=5$ latent dimensions.
The testing instances (T1 and T2 for each category) are not used---neither for training the network nor for building the shape spaces. 
In this manner, the testing instances are completely novel when presented to our approach for evaluation.
Samples of the generated training images are shown in Figure~\ref{fig:generated_imgs}.
Ground truth object poses are used in order to evaluate only the performance of the non-rigid registration.
The robustness of our method is however evaluated against noise on the object pose.
Note that object poses can be estimated using approaches such as~\cite{capellen2019convposecnn} or~\cite{xiang2018posecnn}.

The training images are split up randomly. 
90$\%$ are used for actual training and 10$\%$ are used for validation. 
For training, the ground truth deformation are scaled by a factor of 1000, 
to make a clear boundary between foreground and background.
The $L_2$ loss is used for training the network. 
We trained on two graphic cards with a batch size of 24 each (effective batch size of 48). 
The learning rate is decreased stepwise, starting from $3\times10^{-5}$ it is divided by two every 600 epochs, until reaching a final learning rate of $1\times10^{-6}$. 
The training is done for 2000 epochs. 

\subsection{Experimental Results}
\label{sec:experimental_results}
\begin{figure}[b!]
	\vspace*{-2ex}
	\centering
	\includegraphics[width=\linewidth]{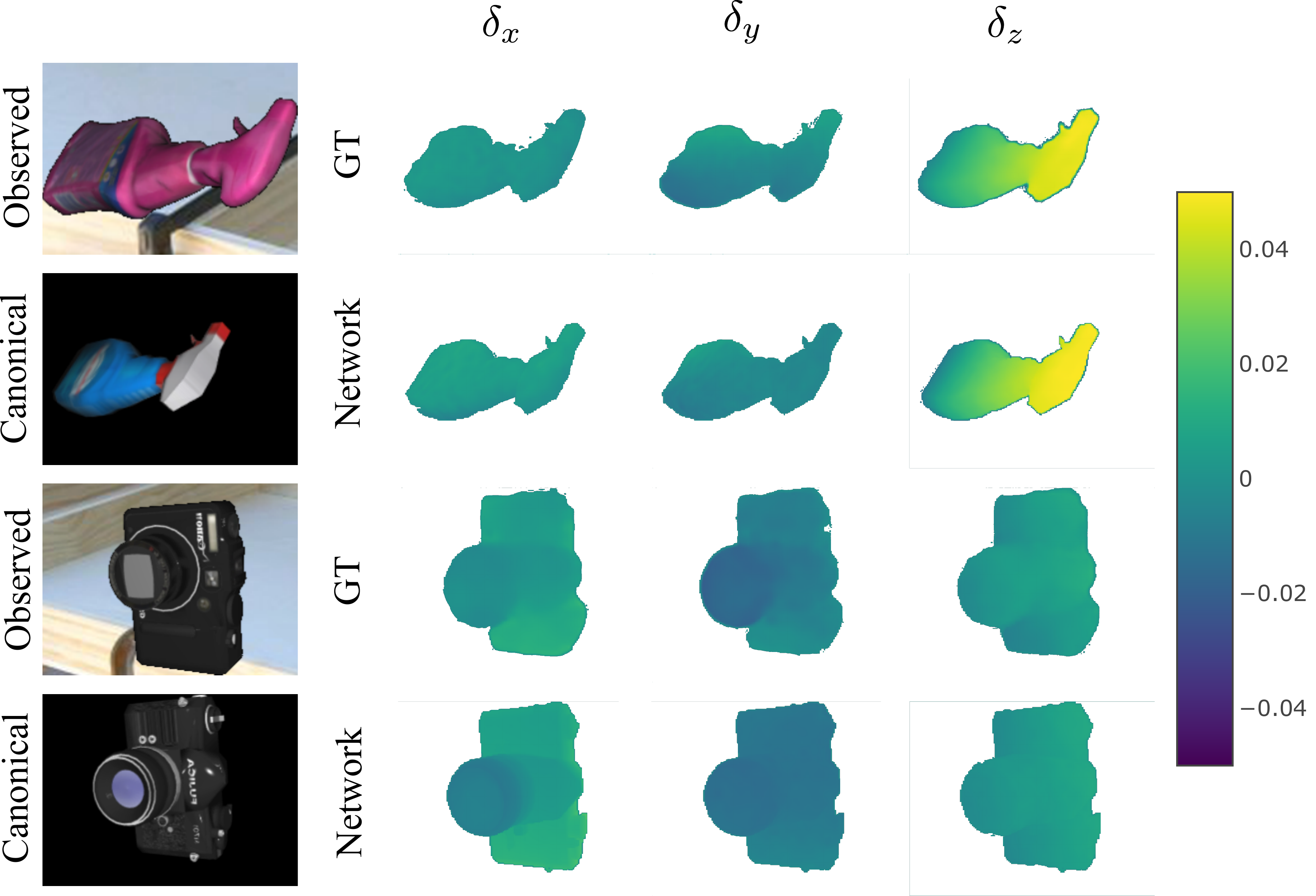}
	\caption{Estimated deformations (in meters) of novel instances presented to the network.
		For comparison, the ground truth (GT) is also shown.
		For clarity in the visualization, background pixels are painted white.	
	}
	\label{fig:gt_nw}
\end{figure}
To evaluate the quality of the registration, 
the following error function between two point clouds $\mathbf{C} \in \mathbb{R}^{n \times 3}$ and $\mathbf{T} \in \mathbb{R}^{m \times 3}$ is defined:
\begin{equation}\label{error}
E(\mathbf{T}, \mathbf{C}) = \frac{1}{m} \sum_{i = 0}^{m - 1} \underset{j}{\min} \left\lVert \mathbf{T}(i,.) - \mathbf{C}(j,.) \right\rVert ^ 2.
\end{equation}
It computes the mean distance between the points of the observed model to the respective closest point of the deformed canonical point cloud. 

We compare our results against the Categorical Latent Space (CLS) approach~\cite{Rodriguez2018a} and CPD~\cite{myronenko2010point}.
Note,
however, 
that those approaches require 3D data in contrast to our method.
The latent space approach~\cite{Rodriguez2018a} and the CPD method are parameterized using the same values as for generating the training images.
This encourages a fair evaluation,
because a bad parametrization of CPD will affect the quality of our training images and shape space.
The latent space dimensionality of CLS is also set to five. 
To generate testing images, we rendered the testing instances from 74 points of view on a tessellated sphere and produced a point cloud to represent the observed parts of the instances. 
\begin{table}[t]
	\centering
	\caption{Comparison with other approaches.\hspace{\textwidth} Mean and (standard deviation) error values expressed in $\mu$m.}
	\vspace*{-1ex}
	\label{table:error}
	\begin{tabular}{lcccc}
		\toprule
		Instance & Ground & CLS~\cite{Rodriguez2018a} & CPD~\cite{myronenko2010point} & Ours \\
		& Truth & (3D data) & (3D data) & (RGB) \\
		\midrule
		Camera T1 & 34.61 & 51.93 & 407.04 & 102.17 \\
		& (1.97) & (10.45) & (491.89) & (47.89) \\
		Camera T2 & 16.45 & 19.87 & 167.18 & 18.80 \\
		& (1.61) & (4.59) & (358.16) & (5.11) \\
		\midrule
		Bottle T1 & 23.25 & 25.92 & 140.51 & 45.21 \\
		& (2.34) & (5.18) & (312.13) & (9.75) \\
		Bottle T2 & 90.42 & 72.33 & 357.98 & 88.35 \\
		& (28.54) & (11.35) & (536.81) & (18.39) \\
		\midrule
		Sp. Bottle T1 & 29.84 & 30.78 & 298.53 & 47.87 \\
		& (1.42) & (1.89) & (409.14) & (12.99) \\
		Sp. Bottle T2 & 111.94 & 121.19 & 376.09 & 154.97 \\
		& (14.29) & (19.16) & (720.73) & (82.34.39) \\
		\midrule
		Drill T1 & 21.18 & 28.86 & 232.88 & 52.71 \\
		& (0.949) & (1.42) & (1319) & (23.54) \\
		Drill T2 & 63.95 & 58.50 & 216.35 & 119.88 \\
		& (5.23) & (21.51) & (566.18) & (107.43) \\
		\bottomrule
	\end{tabular}
	\vspace*{-2ex}
\end{table}
\begin{figure}[b!]
	\centering
	\includegraphics[width=\linewidth]{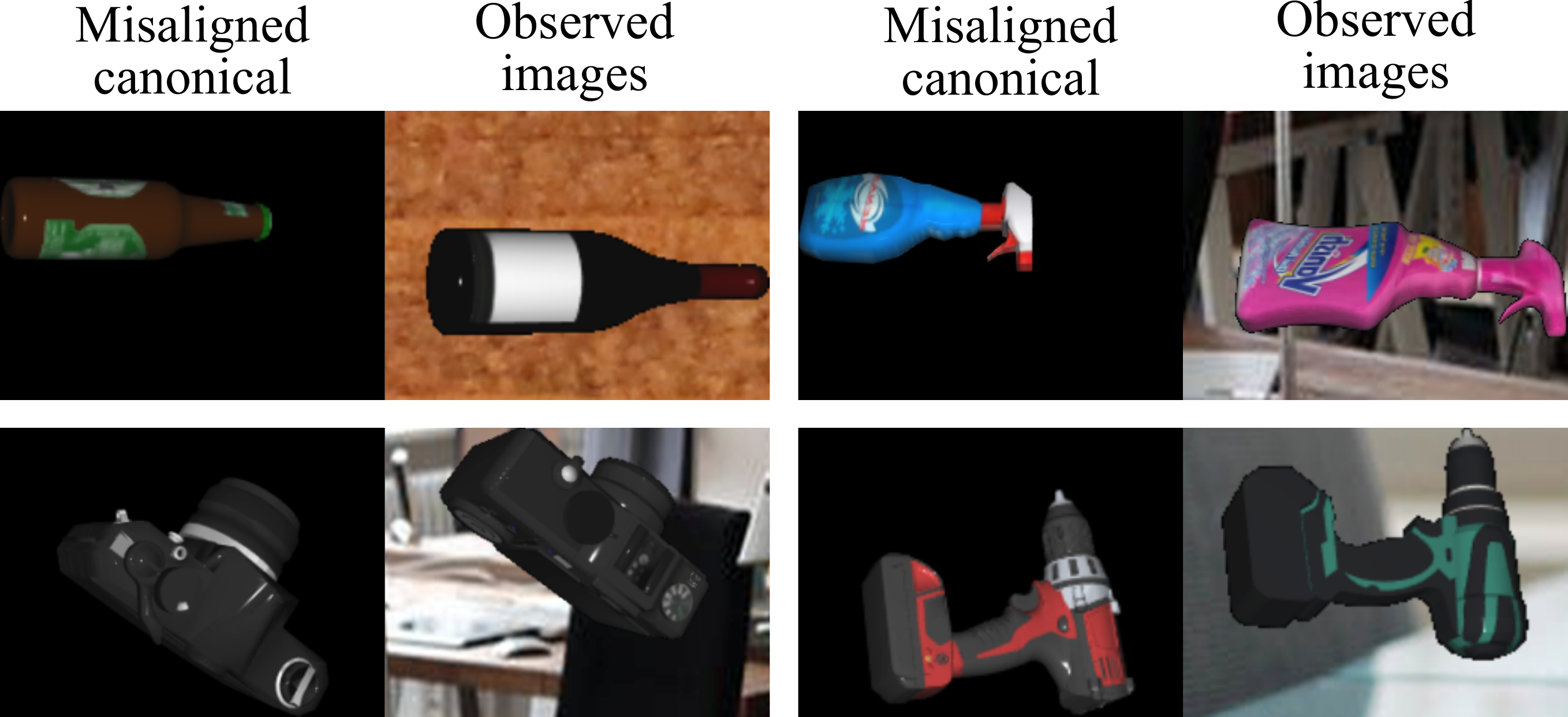}
	\caption{Network input images after noise on the object pose is applied.}
	\label{fig:image_pose_noise}
\end{figure}
In addition, the deformation that results from the ground truth target images is calculated.
This establishes a bound for the results of our approach and allows us to evaluate the network performance.
Figure \ref{fig:gt_nw} shows the ground truth target deformations and the network output for two instances of the spray and camera categories.
The mean and standard deviations of the error values of all four categories are presented in Table~\ref{table:error}.
The CLS approach~\cite{Rodriguez2018a} performs best overall,
since it also incorporates a shape space and had in contrast to our approach access to 3D data.
Our approach outperforms CPD by a large margin even without access to 3D data for the evaluated categories.
Compared to CPD,
the lower error mean indicates that our approach reconstructs the objects better 
while the lower standard deviation demonstrates that out method is more robust to different object viewpoints.

\begin{table}[t]
	\centering
	\caption{Evaluation against noise on the object pose. 
		\hspace{\textwidth}Mean and (standard deviation) error values expressed in $\mu$m.}
	\vspace*{-1ex}
	\label{table:pose_eval}
	\begin{tabular}{lcccc}
		\toprule
		Category & CPD~\cite{myronenko2010point} (3D data) & Ours (RGB) \\
		\midrule
		Camera T1 & 168.54 (357.8) & 105.26 (64.21) \\
		Camera T2  & 406.45 (492.03) & 306.96 (127.89) \\
		\midrule
		Bottle T1 & 297.79 (579.49) & 227.90 (146.0) \\
		Bottle T2 & 852.40 (1818) & 289.36 (147.68) \\
		\midrule
		Spray Bottle T1  & 1035 (406.69) & 146.89 (117.57) \\
		Spray Bottle T2  & 1488 (554.33) & 255.69 (167.32) \\
		\midrule
		Drill T1  & 232.35 (1325) & 92.96 (58.23) \\
		Drill T2  & 215.54 (565.48) & 262.31 (228.40) \\
		\bottomrule
	\end{tabular}
\end{table}

We further evaluated our approach against noise on the given object pose.
A three-dimensional vector is uniformly sampled from $[-0.05, 0.05]$ for each coordinate axis and added to the canonical object pose. 
Resulting rendered images are shown in Figure~\ref{fig:image_pose_noise}.
We compare our performance against CPD; the results are presented in Table~\ref{table:pose_eval}.
As expected,
the performance of our approach is affected by this large misalignment, 
because no pose refinement is explicitly modeled.
However, we achieved lower registration errors compared to CPD.

\begin{figure}[b!]
	\centering
	\includegraphics[width=\linewidth]{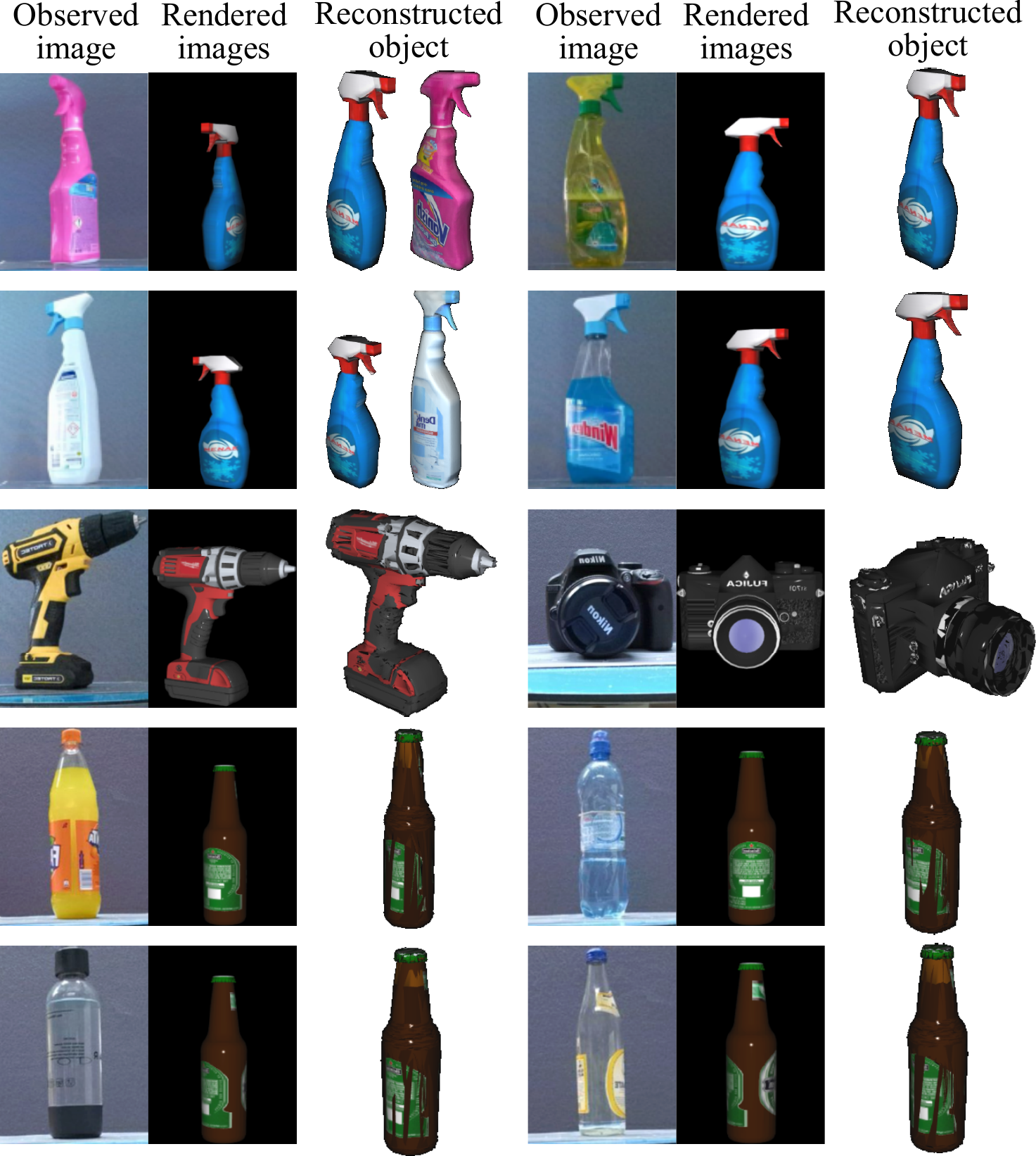}
	\caption{Non-rigid registration on real pictures.
		The ground truth of two objects was available and is presented next to the reconstructed object.
	Hard objects to perceive as transparent bottles are successfully registered.}
	\label{fig:real_images}
\end{figure}
We finally evaluated the performance of our approach with real object images.
Figure~\ref{fig:real_images} shows the input to the network and the resulting deformed models.
The applicability of our method on real images is demonstrated based on the plausible registration results.
The reconstruction of the white spray bottle does not match well with the ground truth possibly because none of the training instances exhibits such large dimensions.
The ground truth 3D models were available for two real spray bottles.
We computed the registration error for 12 different object poses by turning these two real objects around their vertical axis from their standing position every 30$^{\circ}$.
The results are presented in Table~\ref{table:real_eval}.

\begin{table}[t]
	\centering
	\caption{Evaluation with real images. \hspace{\textwidth}Mean and (standard deviation) error values expressed in $\mu$m. }
	\vspace*{-1ex}
	\label{table:real_eval}
	\begin{tabular}{lcccc}
		\toprule
		Category & CPD~\cite{myronenko2010point} (3D data) & Ours (RGB) \\
		\midrule
		Real Spray B. (pink)  & 451.78 (329.7) & 67.34 (41.18)\\
		Real Spray B. (white)  & 436.38 (409.25) & 398.03 (229.05) \\
		\bottomrule
	\end{tabular}
\end{table}

\section{Conclusion}
\label{sec:conclusion}
We presented a novel approach for category-level non-rigid registration based on single-view RGB images.
We demonstrated that a neural network is able to infer deformations on the visible parts of the observed object.
In addition, we showed how objects are reconstructed by incorporating a learned shape space for each category.
We evaluated our approach on synthetic and real images outperforming CPD even without depth information and with noise on the object pose.
Furthermore, we demonstrated that objects which are hard-to-measure by depth sensors (e.g., transparent bottles) can be successfully registered with our method.

In the future, we plan to extend our neural network in order to infer pixel-wise object categories to better match the masks used for training.
Furthermore, we will incorporate an object pose refinement module on the network and the shape space to increase robustness against misalignments.	


\balance
\bibliographystyle{IEEEtranN}
\bibliography{deep_cpd}

\end{document}